\pdfoutput=1

\documentclass[11pt]{article}

\usepackage[preprint]{acl}

\usepackage{times}
\usepackage{latexsym}

\usepackage[T1]{fontenc}

\usepackage[utf8]{inputenc}

\usepackage{microtype}

\usepackage{inconsolata}

\usepackage{graphicx}

\usepackage{amsfonts}
\usepackage{amsmath}
\usepackage{booktabs}
\usepackage{multirow}
\usepackage{subcaption}

\usepackage{graphicx}
\usepackage{algorithm}
\usepackage{algpseudocode}
\algrenewcommand\algorithmicrequire{\textbf{Input:}}
\algrenewcommand\algorithmicensure{\textbf{Output:}}

\usepackage{caption}
\usepackage{bbold}
\usepackage{cleveref}
\crefname{section}{§}{§§}

\usepackage{tikz}

\usepackage{tabularx}

\usepackage{color,soul}
\usepackage{paralist}
\usepackage[shortlabels]{enumitem}
\usepackage{epigraph}

\usepackage[most]{tcolorbox}
\usepackage{tikz}
\newtcolorbox[list inside=prompt,auto counter,number within=section]{prompt}[1][]{
    colbacktitle=black!60,
    fonttitle=\small,
    coltitle=white,
    fontupper=\footnotesize,
    boxsep=4pt,
    left=0pt,
    right=0pt,
    top=0pt,
    bottom=0pt,
    boxrule=1pt,
    #1,
}
\usepackage{etoolbox}

\makeatletter
\patchcmd{\epigraph}{\@epitext{#1}}{\itshape\@epitext{#1}}{}{}
\makeatother

\title{On the Mutual Influence of \\Gender and Occupation in LLM Representations}

\author{Haozhe An~~~~~~Connor Baumler~~~~~~Abhilasha Sancheti~~~~~~Rachel Rudinger \\
    University of Maryland, College Park \\
    \texttt{\{haozhe, baumler, sancheti, rudinger\}@umd.edu} \\
}

\begin{document}
\maketitle
\begin{abstract}
We examine LLM representations of gender for first names in various occupational contexts to study how occupations and the gender perception of first names in LLMs influence each other mutually. We find that LLMs' first-name gender representations correlate with real-world gender statistics associated with the name, and are influenced by the co-occurrence of stereotypically feminine or masculine occupations. Additionally, we study the influence of first-name gender representations on LLMs in a downstream occupation prediction task and their potential as an internal metric to identify extrinsic model biases. While feminine first-name embeddings often raise the probabilities for female-dominated jobs (and vice versa for male-dominated jobs), reliably using these internal gender representations for bias detection remains challenging.
\end{abstract}

\section{Introduction}
Gender-occupation stereotypes have long been a challenge to address in language technology systems~\cite{rudinger-etal-2018-gender, zhao-etal-2018-gender, romanov-etal-2019-whats, sun-etal-2019-mitigating, sheng-etal-2019-woman, ju-etal-2024-female, wan2024male}.
As first names are often used as proxies for gender, society may develop expectations and stereotypes about occupational roles associated with gender-revealing names.
Social science research demonstrates that such stereotypes can cause harm in education~\cite{harari1973name, pozo-2020-whether} and employment~\cite{smith2005name}, as individuals receive disparate treatment based solely on their feminine or masculine first names, even when all other factors are held constant.

With the advent of large language models~\citep[LLMs;][]{achiam2023gpt, team2023gemini, jiang2023mistral, dubey2024llama}, prior studies have shown some of them exhibit human-like biases and leverage gender stereotypes about first names~\cite{eloundou2025firstperson} when making hiring decisions~\cite{an-etal-2024-large, nghiem-etal-2024-gotta, Wilson_Caliskan_2024}, writing recommendation letters~\cite{wan-etal-2023-kelly}, and generating predictions about romantic relationships from dialogues~\cite{sancheti-etal-2024-influence}.
In particular, gender-occupation stereotypes are a well studied research topic~\cite{kotek_2023_gender, veldanda2023investigating, wang-etal-2024-jobfair, leong2024gender, zhang-etal-2025-hire}.
However, existing work mostly takes a black-box approach, providing limited insights into the potential causes of model behavior that mimics human-like gender stereotypes associated with first names, leaving the \textit{why} and \textit{how} as open questions.
We address this gap by examining models' internal representations of gender. While prior work has studied the gender information encoded in embeddings~\cite{bolukbasi2016man, caliskan2017semantics, basta-etal-2019-evaluating}, we establish connections between models' internal gender representations and their biased behavior in a downstream task associated with first names, demonstrating the mutual influence between occupation and LLM representation of gender for first names.

\begin{figure}[t]
	\centering
	\includegraphics[width=0.95\linewidth]{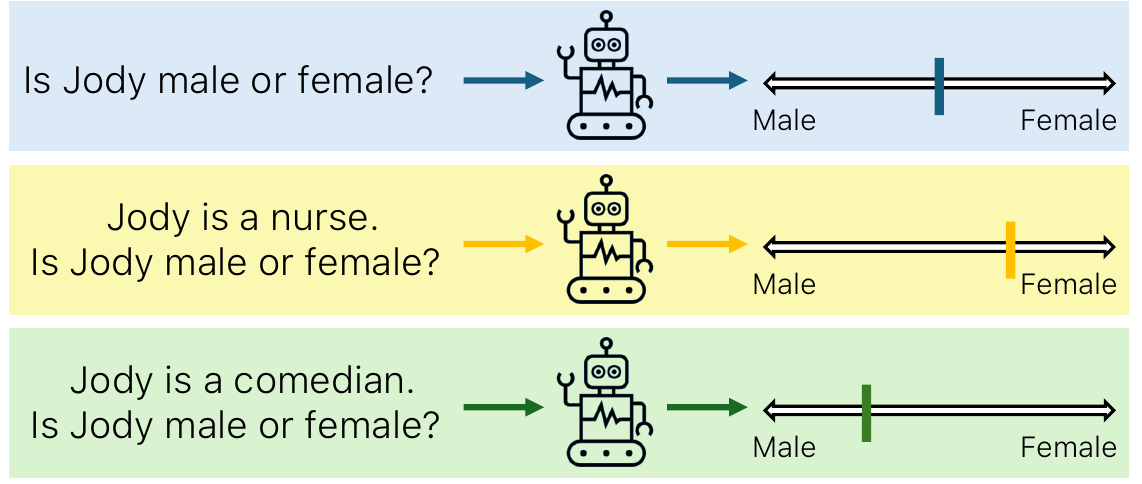} 
 \caption{We derive first-name gender representations in LLMs by projecting their contextualized embeddings onto an approximated gender direction. We find that these representations shift with the occupational context, e.g., ``nurse'' ($90.9\%$ female) increases femininity, while ``comedian'' ($21.1\%$ female) skews masculinity. We also examine how these gender representations correlate with biased behavior in downstream occupation prediction.}
	\label{fig:teaser}
\end{figure}

To explain the model's biased behavior, we conduct a systematic study of the internal gender representations of first names in LLMs across occupational contexts, as illustrated in Fig.~\ref{fig:teaser}.
We obtain the first-name gender representations by projecting their respective contextualized embeddings onto an approximated gender direction, computed by adapting an existing gender direction algorithm~\cite{bolukbasi2016man, basta-etal-2019-evaluating} to open-source LLMs with thorough validation~(\cref{sec:gender_rep_in_llms}).
Our observations show: 
\begin{inparaenum}[(a)] 
\item that these internal representations reveal the gender assumptions made by LLMs about first names, which align with real-world statistics~(\cref{sec:align_rep_and_world}); and 
\item that the gender representations can be sensitive to the contextual information about occupation~(\cref{sec:rep_and_context}). 
\end{inparaenum} 
Furthermore, in the downstream task of occupation prediction from biographies~\cite{dearteaga-2019-biasinbios}, we perform a series of counterfactual name-replacement experiments using a fixed set of contexts (including biographies of both female and male individuals) to isolate the influence of different first names. Our results, based on over 12 million prompts across four LLMs, show that LLMs achieve higher true positive rates for gender-biased occupations when the gender associated with a first name aligns with the bias.
Our analysis highlights the internal trends between first-name representations and occupation prediction in LLMs that may explain this biased behavior, albeit with limited correlation (\cref{sec:downstream}).

While most studies focus on the binary gender association of first names for simplicity~\cite{hall-maudslay-etal-2019-name, wang-etal-2022-measuring, wan-etal-2023-kelly}, our paper enhances the interpretability of LLMs' representations of first names across varying degrees of femininity by including gender-ambiguous names, further enriching the observations from~\citet{you-etal-2024-beyond}. However, we recognize that our analysis does not cover all gender identities.

\section{Related Work}

We situate our work within the literature, connecting our contributions to related studies.

\paragraph{Biases in Embeddings}
Research shows that both static and contextualized embeddings encode human-like biases, including gender and occupation stereotypes~\cite{bolukbasi2016man, zhao-etal-2018-learning, gonen-goldberg-2019-lipstick, may-etal-2019-measuring, zhao-etal-2019-gender, basta-etal-2019-evaluating, dev-etal-2021-oscar, kaneko-bollegala-2021-debiasing, kaneko-etal-2022-gender-bias}. These insights motivate our study of first-name gender representation in LLMs and its link to LLM behavior in a downstream task.

\paragraph{Gender Representation in Embeddings}
\citet{bolukbasi2016man} proposed identifying a gender subspace in static word embeddings, such as Word2Vec~\cite{mikolov2013distributed}, using gendered word pairs (e.g., ``she''–``he'') and defining the gender direction as the top principal component of their embedding differences. This approximation enables projection-based debiasing algorithms~\cite{bolukbasi2016man, dev2019attenuating, wang-etal-2020-double, an-etal-2022-learning} to reduce gender bias. Similar methods have been applied to contextualized embeddings~\cite{peters-etal-2018-deep, devlin-etal-2019-bert} using principal components to approximate the gender subspace~\cite{zhao-etal-2019-gender, basta-etal-2019-evaluating, liang-etal-2020-towards}. We adopt this approach to analyze gender representation in first-name embeddings in LLMs.

\paragraph{First Names and Demographic Attributes}
Despite limitations in associating names with demographic attributes~\cite{gautam-etal-2024-stop}, names are commonly used as proxies for gender, race/ethnicity, and nationality~\cite{greenwald1998measuring, bertrand2004emily, caliskan2017semantics, baumler-rudinger-2022-recognition, an-etal-2023-sodapop, sandoval-etal-2023-rose, acquaye-etal-2024-susu, zhang2024climb}. This usage reflects a strong association between names and demographic factors, both in reality and in model representations. We verify the correlation between LLMs' first-name gender representations and real-world statistics, and explore how these representations vary with context.

\paragraph{Name Artifacts}
In addition to demographic attributes, language models treat names based on factors like frequency~\cite{wolfe-caliskan-2021-low}, tokenization length~\cite{an-rudinger-2023-nichelle}, and associations with prominent entities~\cite{shwartz-etal-2020-grounded}. While we acknowledge these artifacts, this paper focuses on LLM representations of gender for first names and their mutual influence on occupation mentions or predictions.

\begin{figure*}[t]
	\centering
        \begin{subfigure}[]{0.19\linewidth}
		\centering
		\includegraphics[width=\linewidth]{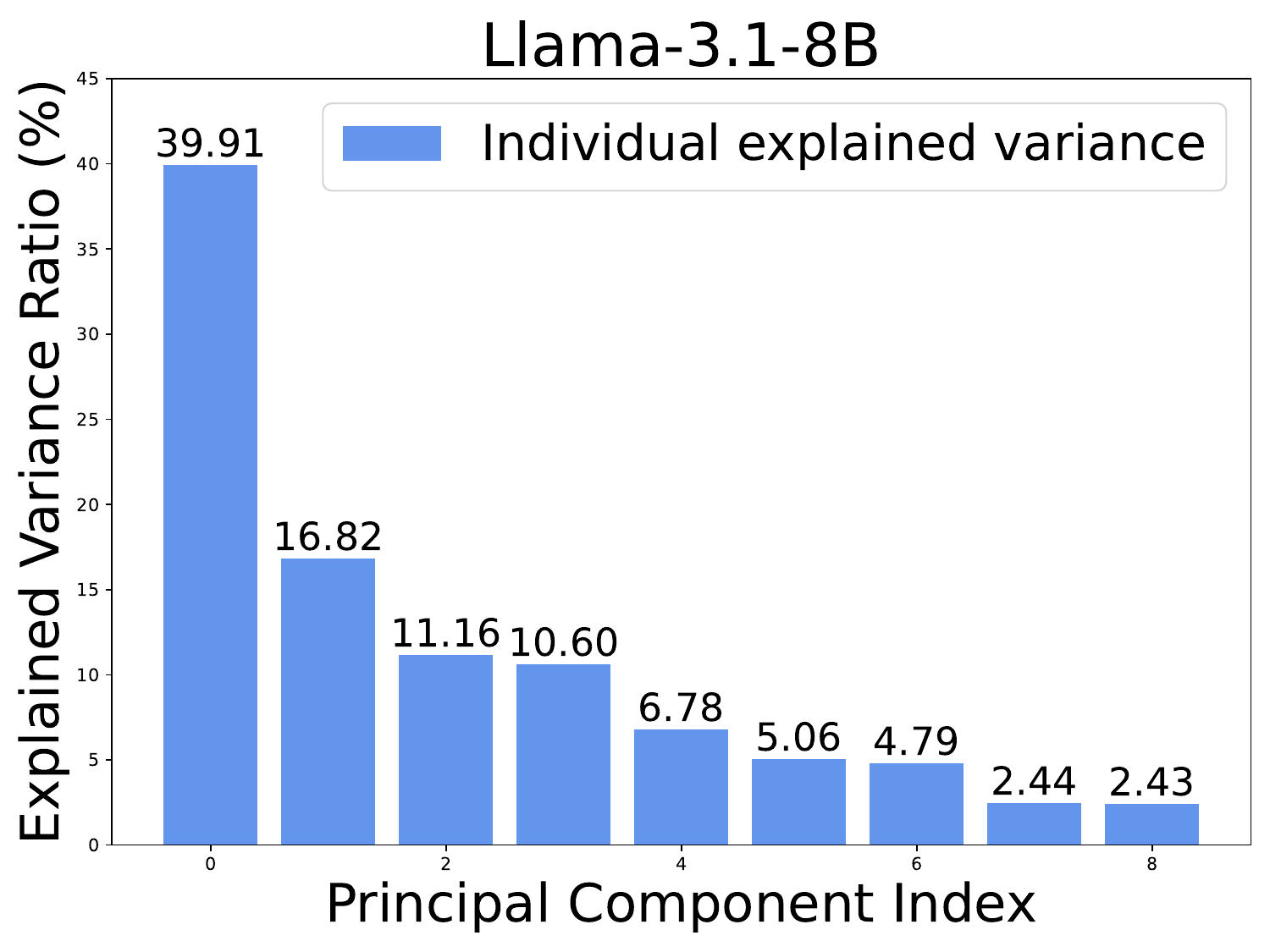}
		\caption{}
		\label{fig:gender_dir_pca_llama}
	\end{subfigure}
	\hfill
        \begin{subfigure}[]{0.19\linewidth}
		\centering
		\includegraphics[width=\linewidth]{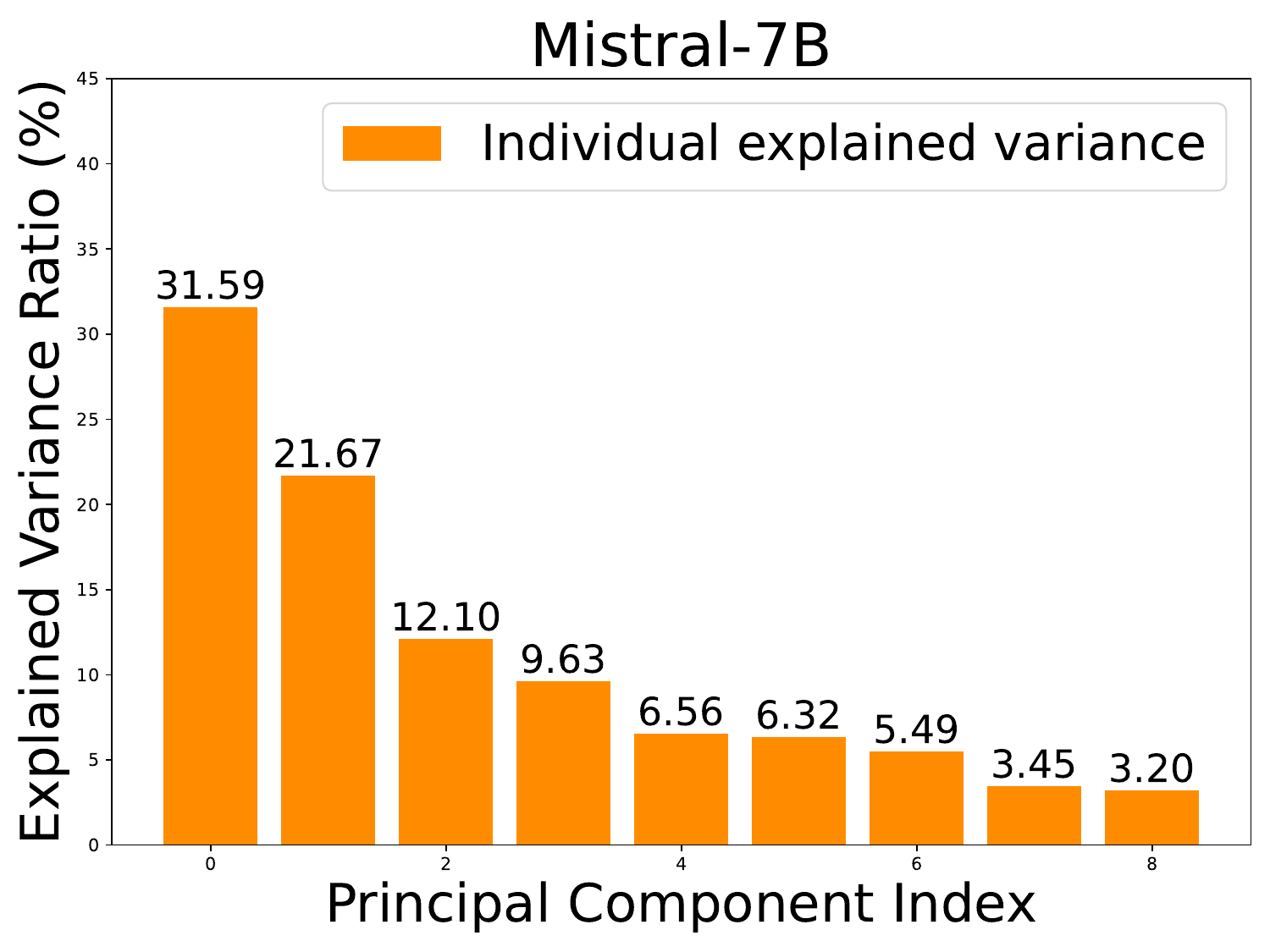}
		\caption{}
		\label{fig:gender_dir_pca_mistral}
	\end{subfigure} 
	\hfill
 	\begin{subfigure}[]{0.19\linewidth}
		\centering
		\includegraphics[width=\linewidth]{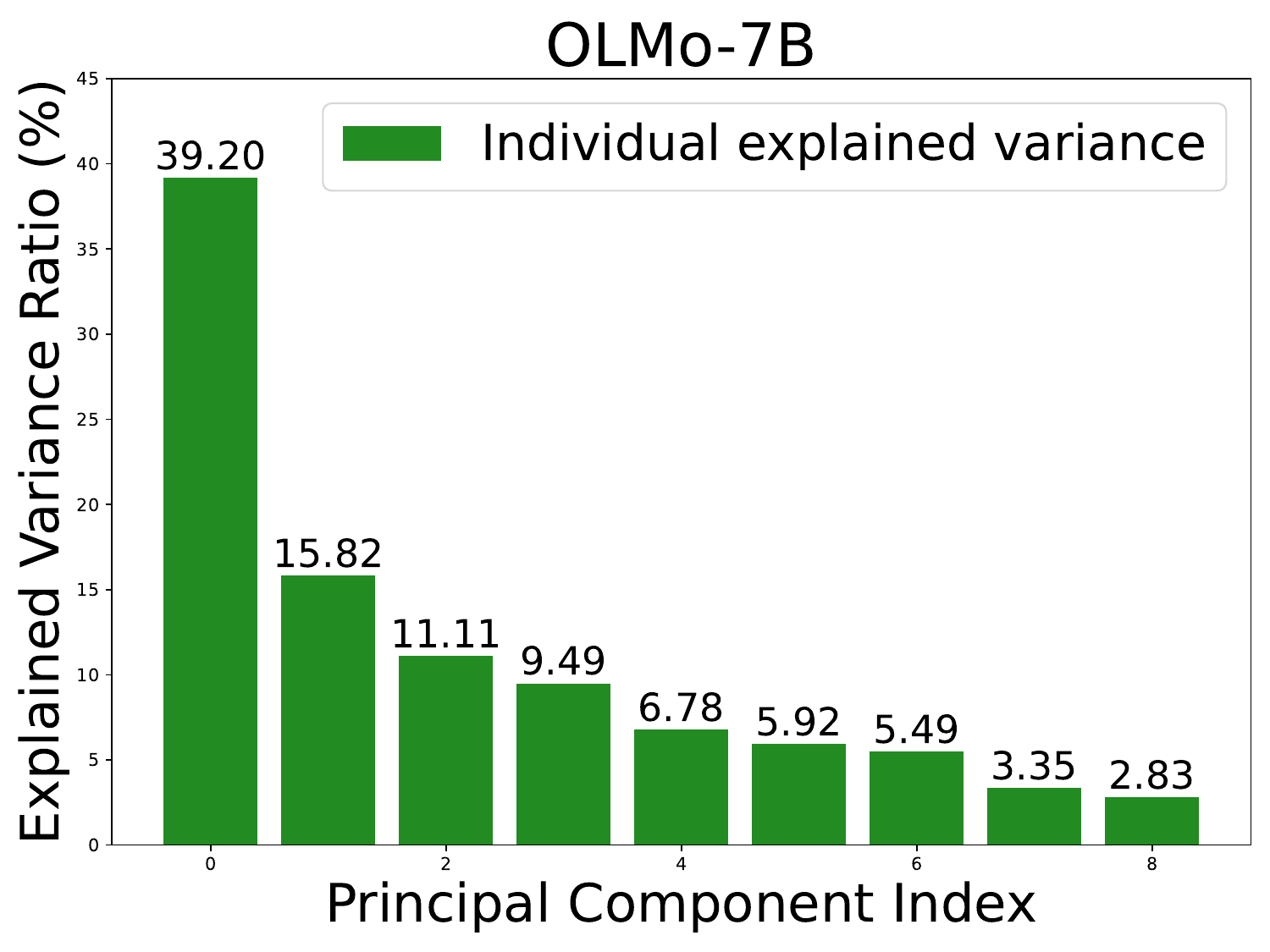}
		\caption{}
		\label{fig:gender_dir_pca_olmo}
	\end{subfigure}
    \hfill
    \begin{subfigure}[]{0.19\linewidth}
		\centering
		\includegraphics[width=\linewidth]{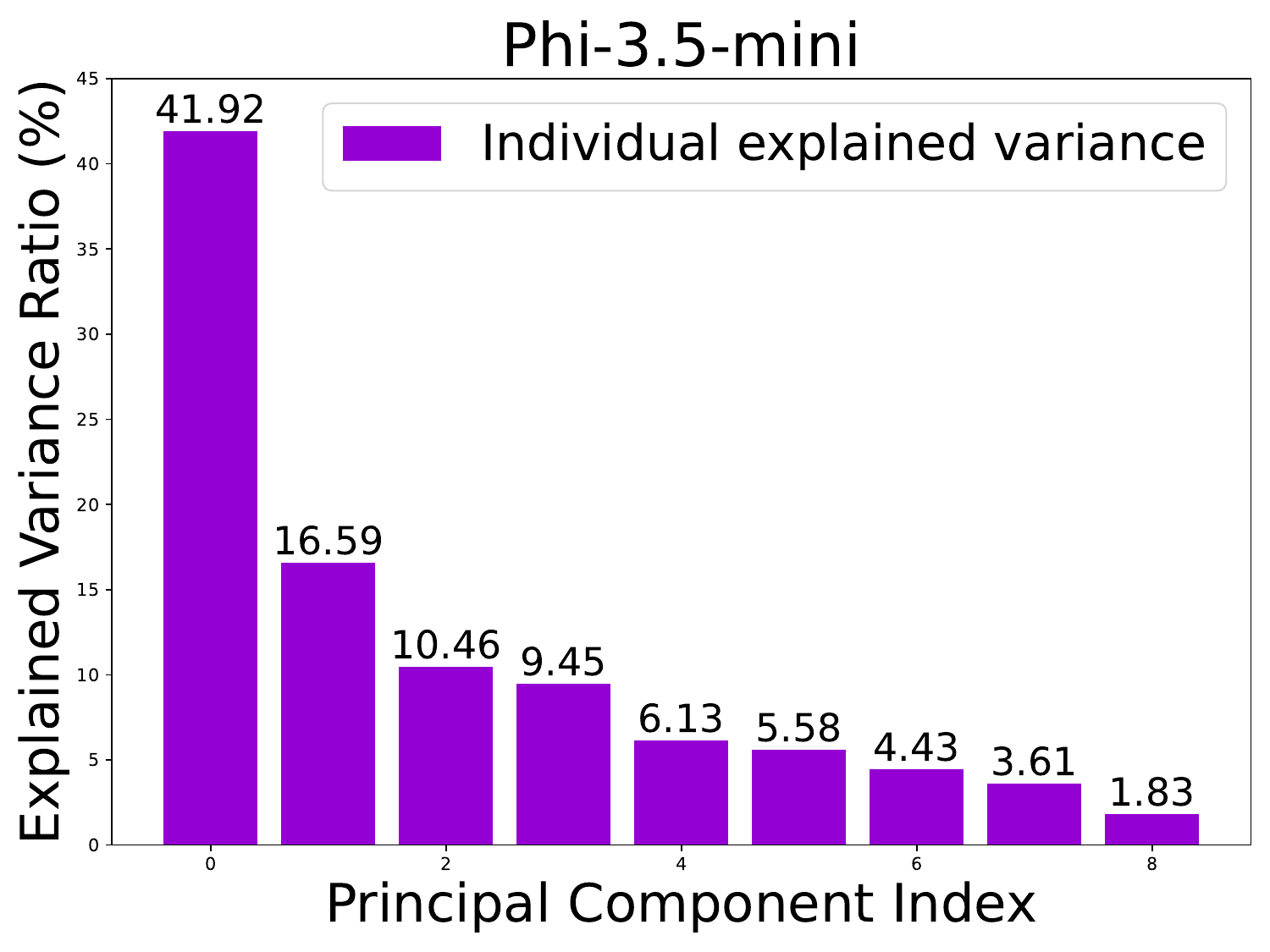}
		\caption{}
		\label{fig:gender_dir_pca_phi}
	\end{subfigure} 
    \hfill
    \begin{subfigure}[]{0.19\linewidth}
		\centering
		\includegraphics[width=\linewidth]{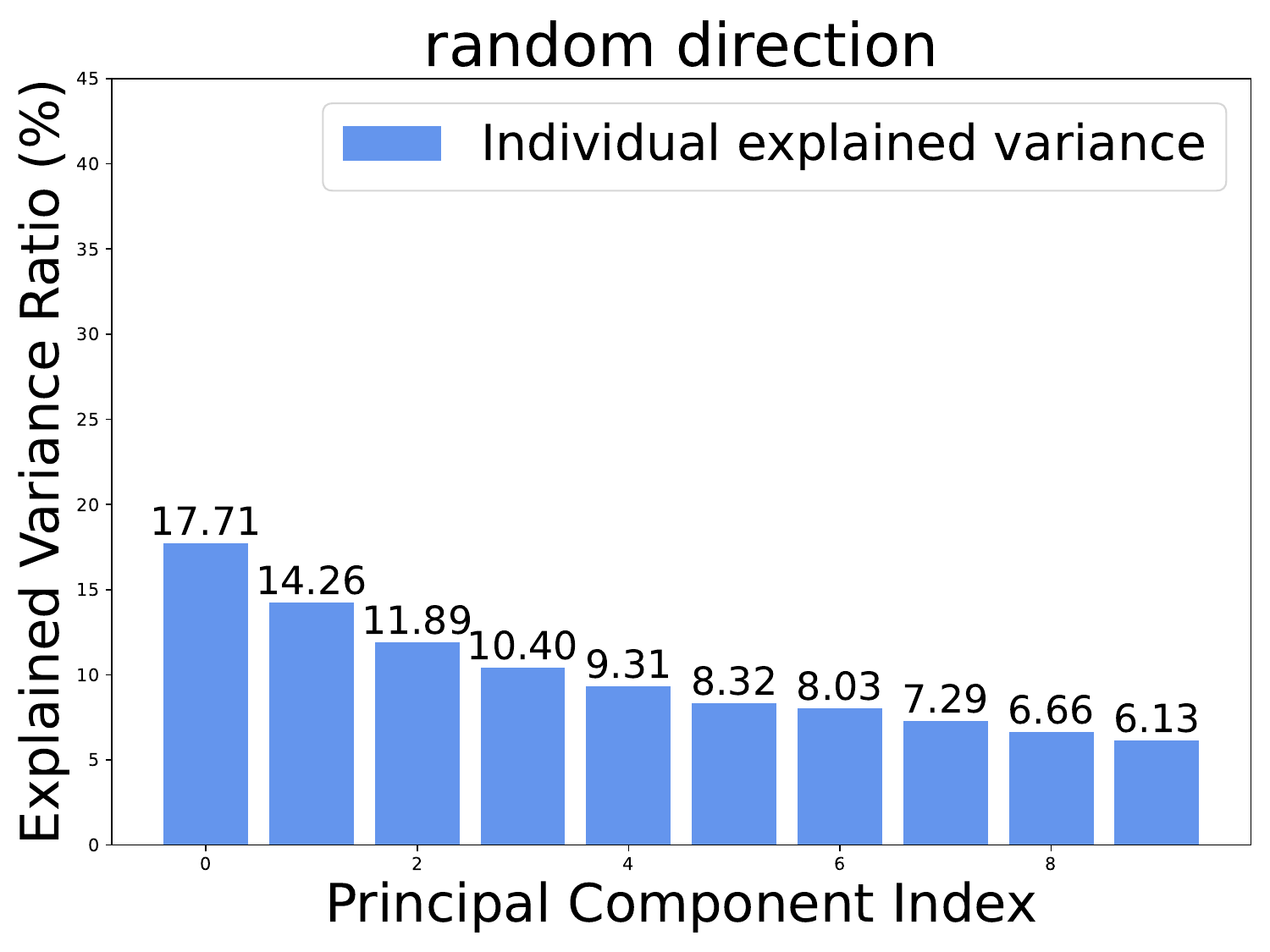}
		\caption{}
		\label{fig:gender_dir_pca_random}
	\end{subfigure} 
    \caption{The percentage of variance explained in the principal components as a result of applying PCA to the differences between gendered (or random) word embeddings from various models. 
    These results indicate that the first PC primarily captures the gender subspace in the respective LLM embedding space.
        }
	\label{fig:gender_dir_pca}
\end{figure*}

\section{Gender Direction in LLMs}
\label{sec:gender_rep_in_llms}

To study gender representation in first names, we derive a vector approximating the female-male direction using an existing gender direction approximation algorithm~\cite{bolukbasi2016man, basta-etal-2019-evaluating, liang-etal-2020-towards}. We evaluate the quality of gender direction approximation, with a binary classification task inspired by~\citet{you-etal-2024-beyond}, predicting the gender associated with first names from their LLM representations. While this task may raise ethical concerns, discussed in detail in the ethical considerations section, it validates the approximated gender direction well captures the gender concept in first name representations.

We examine four open-source LLMs on Hugging Face: \texttt{Llama-3.1-8B-Instruct}~\cite{dubey2024llama}, \texttt{Mistral-7B-Instruct-v0.3}~\cite{jiang2023mistral}, \texttt{OLMo-7B-0724-hf}~\cite{Groeneveld2023OLMo}, and \texttt{Phi-3.5-mini-instruct}~\citep[3.8B;][]{abdin2024phi}.
We select these four LLMs in our paper because they are open-sourced, allowing us to study their internal embeddings. Additionally, theses popular models come from different organizations, which likely represent differing training methodologies. Using these models enables us to demonstrate the generalizability of our findings.

\subsection{Gender Direction Approximation}
\label{sec:gender_dir_llm}
To identify the female-male gender direction $\vec{g}$ in LLMs, we first find a gender subspace $V = \{\vec{v_1}, \vec{v_2}, \dots, \vec{v_k} \}$ with $k$ orthogonal vectors $\vec{v_i}$ obtained from principal component analysis (PCA),

\begin{equation}
    V = \mathrm{PCA}_k\left( \bigcup_{j=1}^d \bigcup_{\vec{w} \in \mathcal{G}_j} (\vec{w} - \vec{c}_j) \right)
\end{equation}
where $\vec{w}$ is the average contextualized embedding of a word from the set of $d$ pairs of gendered words $\mathcal{G}$ ($d=9$ in our implementation), 
and $\mathcal{G}_j$ is the $j$th pair (e.g., ``she'' and ``he''). 
The center of one pair of embeddings is given by $\vec{c}_j = \frac{1}{2}\sum_{\vec{w} \in \mathcal{G}_j} \vec{w}$.
The hyperparameter $k$ is empirically determined based on the explained variance ratios. In our paper, we choose $k=1$ and obtain a gender direction.

\paragraph{Gendered Words $\mathcal{G}$}
We use the same set of gendered words as~\citet{bolukbasi2016man}, excluding ``Mary'' and ``John.''
This exclusion is intended to avoid using first names in gender direction approximation, thereby minimizing the risk of overfitting in our first-name representation analysis.
Table~\ref{tab:def_words} in the appendix displays these gendered words.

\paragraph{Average Embeddings $\vec{w}$}
Following~\citet{basta-etal-2019-evaluating}, we compute the average contextualized embeddings of gendered words. From the English Wikipedia corpus,\footnote{\url{https://huggingface.co/datasets/wikimedia/wikipedia/tree/main/20231101.en}} we extract $3,000$ sentences containing each gendered word and create counterfactuals by swapping every word with its counterpart. 
For repeated gendered words $w_1, w_2, \dots, w_n$ in a sentence, embeddings of $w_i$ are averaged as 
$ \vec{w} = \frac{1}{n} \sum_{i=1}^n \vec{w}_i.$ 
If a word $w_i$ is multiply tokenized in the form as $w_i = (t_1, t_2, \dots, t_m)$, then 
$\vec{w_i} = \frac{1}{m} \sum_{j=1}^m  \vec{t}_j.$ 
Finally, we average embeddings across $6,000$ contexts and compute pairwise differences to obtain the difference matrix.

\paragraph{PCA Results}
Upon applying PCA to the difference matrix, we find that the first principal component (PC) explains a significantly higher percentage of variance compared to the others, and this trend holds across all four models in our study, as shown in Fig.~\ref{fig:gender_dir_pca_llama} through Fig.~\ref{fig:gender_dir_pca_phi}. 
In contrast, the PCA results of a random difference matrix obtained from 10 pairs of random words (Table~\ref{tab:def_words} in~\cref{sec:app_gendered_words}) from \texttt{Llama-3.1-8B} show a more gradual change in the percentage variance explained across the PCs, as shown in Fig.~\ref{fig:gender_dir_pca_random}.
Hence, we reach a similar conclusion as~\citet{bolukbasi2016man} and~\citet{basta-etal-2019-evaluating} that the first PC mostly captures the gender subspace in the embedding space of recent LLMs.
For \texttt{Mistral-7B}, however, the notion of gender seems to be better captured with the top two PCs.
Next, we need a method to determine which PCs to include as the approximated gender direction.

\subsection{Gender Direction Evaluation}
\label{sec:gender_dir_eval}
The fact that the first principal component (PC) for each model explains a relatively large proportion of variance (ranging from $32-42\%$ in Fig.~\ref{fig:gender_dir_pca}) implies that most information about gender is contained in the subspace corresponding to the first PC. However, we directly measure the gender information captured in these subspaces via a binary gender prediction task. A good gender direction approximation should preserve the female-male gender information encoded in the original embeddings.

We train classification models using either
\begin{inparaenum}[(1)]
\item the contextualized first-name embeddings or
\item their dot product with the approximated gender direction.
\end{inparaenum}
We collect first names with varying femininity levels and compute their average contextualized embeddings through counterfactual substitution in a fixed set contexts. Then, we compare classification performance using these embeddings versus their projections onto the gender direction to assess the quality of the gender direction.

We note that the purpose of this binary classification task is \textit{not} to predict the gender identities of individuals in the real world. Rather, the purpose is to confirm that any gender information already present in a model's contextualized embeddings of first names is preserved after projection onto the extracted one-dimensional subspace. 
In other words, the classification task serves as validation that the learned subspace is a reasonable proxy for a model's internal representation of gender, enabling the subsequent analyses we perform in this paper.

\paragraph{First Names}
Following~\citet{sancheti-etal-2024-influence}, we sample first names associated with two genders (female and male) and four races/ethnicities (White, Black, Hispanic, and Asian) from the Social Security Application\footnote{\url{https://www.ssa.gov/oact/babynames/}}~(SSA) dataset and a U.S. voter registration dataset~\cite{rosenman2023race}. We select $470$ names with varying degrees of femininity based on the percentage of the female population linked to each name in the SSA dataset. Names are categorized into 10 buckets according to their female distribution percentages (Table~\ref{tab:name_distribution} in appendix). Due to high thresholds for race/ethnicity distribution ($90\%$) and frequency (200), fewer gender-ambiguous names are sampled. 
These first names are also used in the subsequent analysis.
For the binary classification, we label the names as either ``Female'' or ``Male'' based on a $50\%$ threshold.

\paragraph{Contextualized First-Name Embeddings from Wikipedia Contexts}
To minimize contextual influence in obtaining first-name embeddings, we compute average name representations over a fixed set of contexts. We randomly select $24$ names (see~\cref{sec:app_first_names}), evenly distributed across four races/ethnicities and two genders, and retrieve $10$ sentences for each name from the English Wikipedia corpus. Average contextualized embeddings are obtained by counterfactually replacing the original name with each of the $470$ first names.

We use $\vec{n}_{\text{wiki}}$ to denote a contextualized embedding for a first name obtained in this setup.
These embeddings, or their dot product with the gender direction $\vec{g}$, are used as input features to the binary classification models.
We use $70\%$ of the embeddings for training and the remaining $30\%$ for validation, both sampled evenly across demographics.

\begin{table*}[t]
\centering
\scriptsize
\resizebox{\linewidth}{!}{
   \begin{tabular}{@{}lccccc@{}}
\toprule
                                           & \multicolumn{5}{c}{Logistic Regression}                                                                                                        \\ 
\multicolumn{1}{l}{}                      & \multicolumn{1}{c|}{$\vec{n}_{\text{wiki}}$}      & DOT ($\vec{n}_{\text{wiki}}$, $\vec{g}_{\text{1st}}$) & DOT ($\vec{n}_{\text{wiki}}$, $\vec{g}_{\text{2nd}}$) & \multicolumn{1}{c|}{DOT ($\vec{n}_{\text{wiki}}$, $\vec{g}_{\text{avg}}$)} & DOT ($\vec{n}_{wiki}$, random) \\ \midrule
Llama-3.1-8B          & \multicolumn{1}{l|}{\textbf{75.46 $\pm$ 3.19}} & 75.18 $\pm$ 2.62        & 50.78 $\pm$ 5.47         & \multicolumn{1}{l|}{63.97 $\pm$ 3.28}      & 47.09 $\pm$ 3.85    \\
Mistral-7B            & \multicolumn{1}{l|}{\textbf{74.04 $\pm$ 1.46}} & 67.80 $\pm$ 2.18        & 58.44 $\pm$ 2.56         & \multicolumn{1}{l|}{51.77 $\pm$ 2.01}      & 55.18 $\pm$ 1.82    \\
OLMo-7B               & \multicolumn{1}{l|}{76.60 $\pm$ 1.10} & \textbf{80.57 $\pm$ 1.32}        & 55.46 $\pm$ 3.89         & \multicolumn{1}{l|}{63.26 $\pm$ 2.67}      & 56.03 $\pm$ 2.10    \\
Phi-3.5-mini & \multicolumn{1}{l|}{65.67 $\pm$ 4.86} & \textbf{70.64 $\pm$ 1.83}        & 49.08 $\pm$ 1.98         & \multicolumn{1}{l|}{55.60 $\pm$ 3.37}      & 55.60 $\pm$ 2.04    \\ \midrule \midrule
                                           & \multicolumn{5}{c}{Naive Bayes}                                                                                                                \\ 
\multicolumn{1}{l}{}                      & \multicolumn{1}{c|}{$\vec{n}_{wiki}$}      & DOT ($\vec{n}_{\text{wiki}}$, $\vec{g}_{\text{1st}}$) & DOT ($\vec{n}_{\text{wiki}}$, $\vec{g}_{\text{2nd}}$) & \multicolumn{1}{c|}{DOT ($\vec{n}_{\text{wiki}}$, $\vec{g}_{\text{avg}}$)} & DOT ($\vec{n}_{\text{wiki}}$, random) \\ \midrule
Llama-3.1-8B          & \multicolumn{1}{l|}{73.62 $\pm$ 2.74} & \textbf{74.33 $\pm$ 2.63}        & 51.49 $\pm$ 1.39         & \multicolumn{1}{l|}{62.84 $\pm$ 2.27}      & 49.65 $\pm$ 3.50    \\
Mistral-7B            & \multicolumn{1}{l|}{\textbf{71.21 $\pm$ 1.83}} & 64.82 $\pm$ 1.32        & 50.92 $\pm$ 1.82         & \multicolumn{1}{l|}{51.35 $\pm$ 3.00}      & 55.18 $\pm$ 1.13    \\
OLMo-7B               & \multicolumn{1}{l|}{71.91 $\pm$ 1.65} & \textbf{80.28 $\pm$ 1.92}        & 56.03 $\pm$ 3.23         & \multicolumn{1}{l|}{65.11 $\pm$ 4.40}      & 59.43 $\pm$ 4.43    \\
Phi-3.5-mini & \multicolumn{1}{l|}{61.99 $\pm$ 3.25} & \textbf{67.94 $\pm$ 2.17}        & 51.63 $\pm$ 1.38         & \multicolumn{1}{l|}{57.02 $\pm$ 2.27}      & 54.18 $\pm$ 2.79    \\ \bottomrule
\end{tabular}
    }
    \caption{Accuracy (\%) and standard deviation of two classification models for the task of binary gender prediction using various features from the contextualized first-name embeddings. Results are averaged over five runs with different random train-validation splits, with the best accuracy highlighted in bold.
    Using the dot product between the first-name embedding and the first principal component ($\vec{g}_{\text{1st}}$) as input features yields comparable or even improved performance in this task.
    This observation suggests that $\vec{g}_{\text{1st}}$ is a good gender direction approximation.
        }
	\label{tab:gen_dir_binary}
\end{table*}

\paragraph{Binary Gender Classification}
In this task, we train classification models (logistic regression and Naive Bayes respectively) to predict the gender associated with a first name. We consider two types of input features.
The first setup uses the contextualized embedding of a first name from the Wikipedia contexts as the input feature.
This setup serves as a baseline in the evaluation of gender direction approximation quality.
The second setup uses the dot product between the contextualized first-name embedding and the gender direction $\vec{g}$ approximated using the algorithm in~\cref{sec:gender_dir_llm} as the input feature. 
We note that the dot product between two vectors, $\vec{u} \cdot \vec{v}$, is linearly correlated with the projection of $\vec{u}$ onto $\vec{v}$, resulting in equivalent or highly similar feature representations for the classification task.
We hypothesize that the binary gender prediction accuracy would be similar in both setups if the approximated gender direction effectively captures the concept of gender in first name representations.

\paragraph{Combinations of PCs}
We consider three combinations of principal components (PCs) as the approximation of the gender direction.
\begin{inparaenum}[(1)]
    \item $\vec{g}_{\text{1st}}$: Gender direction is represented by the first PC corresponding to the largest variance explained ratio.
    \item $\vec{g}_{\text{2nd}}$: Gender direction is represented by the second PC corresponding to the second largest variance explained ratio.
    \item $\vec{g}_{\text{avg}}$: Gender direction is represented by the average of the first two PCs.
\end{inparaenum}

\paragraph{Evaluation Results}
We report the binary classification accuracy in Table.~\ref{tab:gen_dir_binary}.
We observe that, in line with prior studies~\cite{an-etal-2023-sodapop, sancheti-etal-2024-influence, you-etal-2024-beyond}, the original first-name embeddings $\vec{n}_{\text{wiki}}$ effectively indicate the stereotypical gender associated with the names. This is evidenced by all setups using $\vec{n}_{\text{wiki}}$ as input achieving approximately $60-75\%$ accuracy, which is significantly higher than the random baseline.

Using the dot product between $\vec{n}_{\text{wiki}}$ and $\vec{g}_{\text{1st}}$ as input largely maintains or improves classification accuracy, indicating $\vec{g}_{\text{1st}}$ effectively conveys the gender direction for first-name embeddings. 
In contrast, the second PC and the average of two PCs fail to preserve accuracy, suggesting they are ineffective approximations of the gender direction.

We note that neither PC improves performance over the original setup for \texttt{Mistral-7B}, which aligns with the observation shown in Fig.~\ref{fig:gender_dir_pca_mistral}, where the top PC does not exhibit a notably high variance explained ratio. 
Nonetheless, $\vec{g}_{\text{1st}}$ generally preserves the models' performance for \texttt{Mistral-7B} name embeddings. 
For consistency, we choose to use the first PC $\vec{g}_{\text{1st}}$ as the gender direction approximation $\vec{g}$ for all LLMs in the following analysis.

\section{LLM Representations of Gender in First Name Embeddings}

\begin{figure*}[t]
	\centering
        \begin{subfigure}[]{0.49\linewidth}
		\centering
		\includegraphics[width=\linewidth]{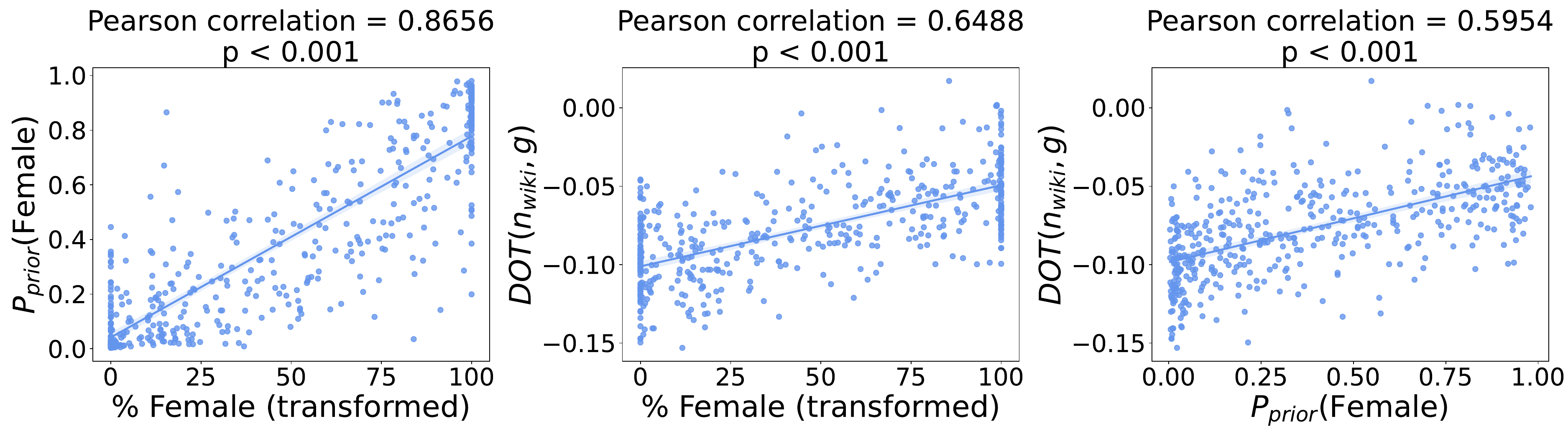}
		\caption{Llama-3.1-8B}
		\label{fig:pearsonr_llama}
	\end{subfigure}
	\hfill
        \begin{subfigure}[]{0.49\linewidth}
		\centering
		\includegraphics[width=\linewidth]{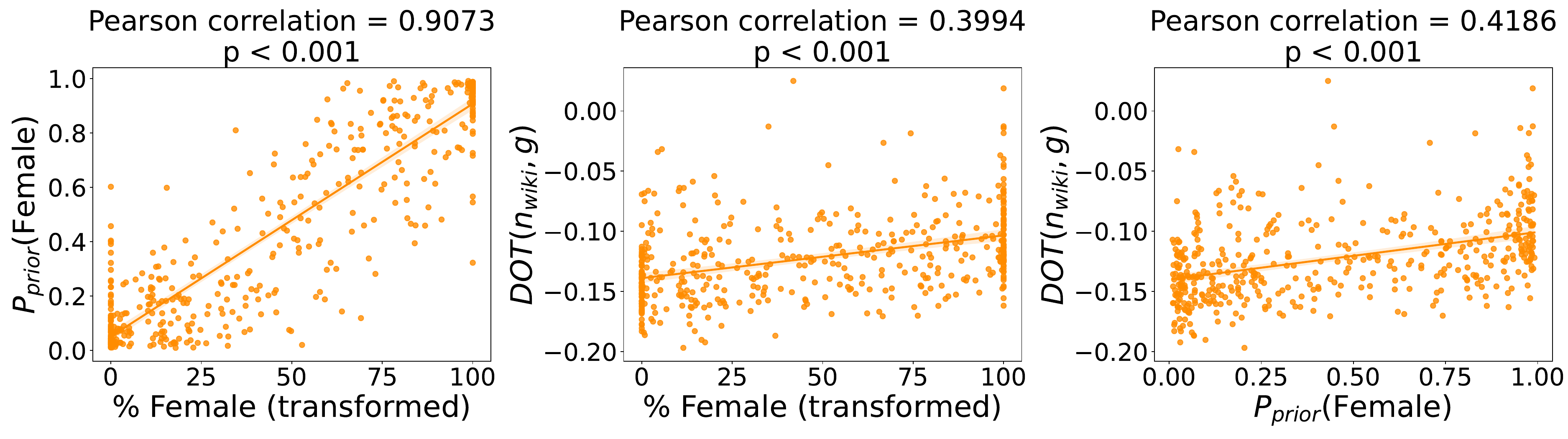}
		\caption{Mistral-7B}
		\label{fig:pearsonr_mistral}
	\end{subfigure} 
    \caption{Scatter plot between each pair of the three variables studied in~\cref{sec:align_rep_and_world} and their Pearson correlation.
    We observe statistically significant linear correlations between each pair of the variables studied. Both the model's prior gender probability and the embedding associated with a name reflect the real-world gender distribution.
    }
	\label{fig:pearsonr_3variables}
\end{figure*}

With a validated gender direction, we show a correlation between the model's representation of gender associated with a name and its real-world gender distribution. We then examine how gender representation varies with mentions of stereotypically feminine or masculine occupations in the context.

\subsection{Correlating LLM Representations of Gender with Real-World Statistics}
\label{sec:align_rep_and_world}

We hypothesize that the model encodes the perceived gender of a first name in alignment with real-world gender distributions, as a result of its training on natural corpora that mirror these correlations.
To validate this hypothesis, we study the correlations between three variables. 

\paragraph{\% Female in Real-World Distribution}
We use the gender distribution associated with first names in the SSA dataset to calculate the percentage of the female population for each first name. 
Because there are more strongly gender-indicative names than gender-neutral names (i.e., more names in the $0-2\%$ bucket than in the $25-50\%$ bucket, shown in Table~\ref{tab:name_distribution} in appendix), we transform each bucket into one of ten linearly divided buckets (e.g., $0-2\%$ is mapped to $0-10\%$, $2-5\%$ is mapped to $10-20\%$, and so on) to smooth the data distribution.

\paragraph{DOT($\vec{n}_{\text{wiki}}, \vec{g}$)}
We reuse the contextualized first-name embeddings $\vec{n}_{\text{wiki}}$ obtained from the set of Wikipedia sentences (\cref{sec:gender_dir_eval}) and compute their dot product with the gender direction $\vec{g}$.
This quantity indicates the degree of femininity (or masculinity) of a first name in the model's representation.

\paragraph{$P_{\text{prior}}(\text{Female})$}
To obtain the gender probability from the actual response of an LLM, we prompt the model with the context \textit{``Question: Is \{NAME\} male or female? Answer: {NAME} is ''} and retrieve the logits for the tokens ``male'' and ``female,'' respectively. The ``\{NAME\}'' placeholder is instantiated with each of the $470$ first names.
We take the softmax of the two gender logits and use the probability for ``female'' as the model's prior gender probability of a name, denoted as $P_{\text{prior}}(\text{Female})$.

\paragraph{Observations}
In Fig.~\ref{fig:pearsonr_3variables} and Fig.~\ref{fig:app_pearsonr_3variables} (appendix), we observe strong (and statistically significant) linear correlations between each pair of the three variables for all LLMs studied. 
We verify that both the model's prior probability of the gender associated with a name and its first-name embeddings reflect the real-world gender distribution.
Furthermore, the prior gender probability also linearly correlates with the degree of femininity in the first-name embeddings. 
We note that, without loss of generality, all gender directions in the four LLMs have been aligned to use the positive direction to denote the female gender for intuitive visualization.

\begin{figure*}[t]
	\centering
        \begin{subfigure}[]{0.49\linewidth}
		\centering
		\includegraphics[width=\linewidth]{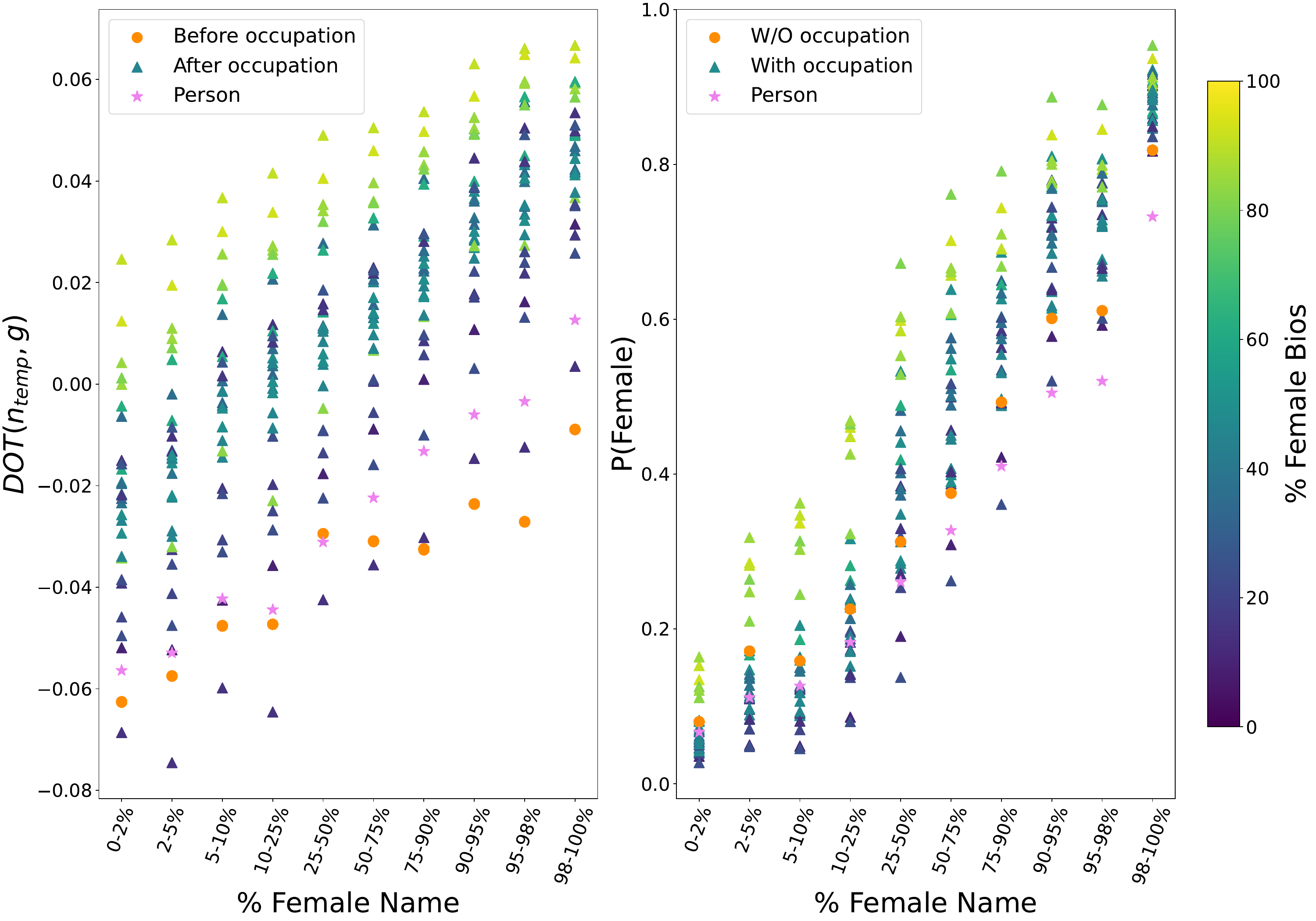}
		\caption{Llama-3.1-8B}
		\label{fig:template_llama}
	\end{subfigure}
	\hfill
        \begin{subfigure}[]{0.49\linewidth}
		\centering
		\includegraphics[width=\linewidth]{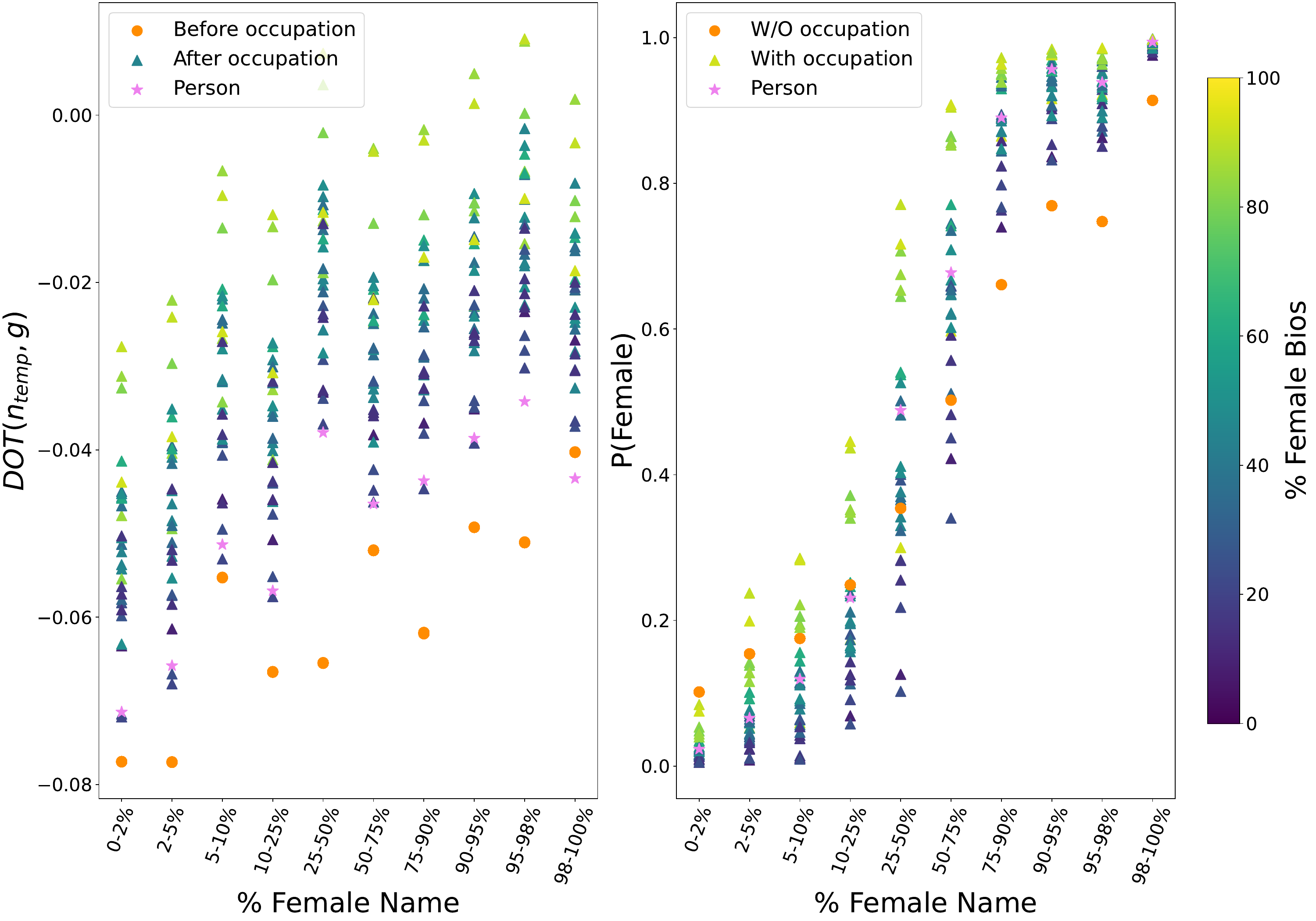}
		\caption{Mistral-7B}
		\label{fig:template_mistral}
	\end{subfigure} \\
    \caption{ \textbf{(Left of each subfigure)} Change of the dot product between the name embedding from a template sentence $\vec{n}_{\text{temp}}$ and the gender direction $\vec{g}$ before and after the mention of an occupation.
    \textbf{(Right of each subfigure)} Change of the output probability of the token ``female'' with and without mentioning an occupation.
    ``\% Female Name'' is the real-world gender distribution of a name (\cref{sec:align_rep_and_world}). 
    ``\% Female Bios'' is the percentage of biographies of female individuals in~\citet{dearteaga-2019-biasinbios}, which mirrors the gender breakdown of an occupation in real life.
    The violet star indicates the non-stereotypical baseline where the occupation placeholder is replaced with the string ``person.''
    We observe that the gender representation of first names generally shifts with occupational contexts, where, within each gender bucket along the horizontal axis, stereotypically female jobs lead to a more positive dot product along the gender direction and a higher predicted probability for the female gender. We also see that the results for strongly masculine or feminine names are less affected by occupation than those for gender-ambiguous names.}
	\label{fig:biosoccupation_template}
\end{figure*}

\subsection{LLM Representations of Gender are Influenced by Occupational Contexts}
\label{sec:rep_and_context}

We investigate whether a model's gender representation of a first name changes under the influence of different occupational contexts.

\begin{prompt}[title={Prompt \thetcbcounter: Gender Prediction}, label=prompt:template]
\scriptsize
Question: \{NAME\} is \{ARTICLE\} \{OCC.\}. Is \{NAME\} male or female? \\
    Answer: \{NAME\} is $\dotsc$
\end{prompt}

\paragraph{Setup}
To investigate whether the model's gender probability changes from $P_{\text{prior}}(\text{Female})$ when an occupation is mentioned, 
we construct sentences using the prompt template~\ref{prompt:template}.
The placeholders 
are replaced with a first name, `a'' or `an,'' and an actual occupation, respectively.
We then retrieve the logits for ``male'' and ``female'' tokens, converting them to probabilities via a softmax function. 
This design choice of converting a subset of token logits to probabilities is inspired by~\citet{duarte-2024-decop}. This approach allows us to capture a continuous distribution of gender probabilities for different first names for our following analysis.

\paragraph{Occupations}
We use the 28 occupations with varying gender dominance from Bias in Bios~\cite{dearteaga-2019-biasinbios}. 
Gender domination of occupations are approximated by the percentage of female biographies in Bias in Bios, which is constructed by scraping real-world biographies that reflect the gender breakdown of an occupation. 
In addition, we introduce another non-stereotypical baseline~\cite{belem2024are}, in which no gender-related language (i.e., stereotypically female or male occupations) is present, in order to illustrate how occupational contexts cause changes in the gender representation of first names. This non-stereotypical baseline is implemented by using the string ``person'' to fill the occupation placeholder.

\paragraph{Metrics}
We compute each first name's gender probability obtained from the prompt containing an occupation (including the non-stereotypical baseline ``person'') and compare it with $P_{\text{prior}}(\text{Female})$ to measure changes in the model's gender perception.
Meanwhile, we also retrieve the first-name embeddings $\vec{n}_{\text{temp}}$ before and after ``\{OCC.\}'' (i.e., the embeddings for the first and second occurrence of ``\{NAME\}''). We find their respective dot product with the gender direction, DOT($\vec{n}_{\text{temp}}, \vec{g}$), and analyze the change between them.
We expect to see a more feminine representation of a first name (higher probability for the ``female'' token and more positive dot product with $\vec{g}$) if the occupation in the context is female-dominated, and vice versa.

\paragraph{LLM Representations of Gender Shift with Contexts}

We present DOT($\vec{n}_{\text{temp}}, \vec{g}$) before and after the occupation mention and $P_{\text{prior}}(\text{Female})$ with and without the occupation in Fig.~\ref{fig:biosoccupation_template}. 
We show the similar visualizations for \texttt{OLMo-7B} and \texttt{Phi-3.5-mini} in Fig.~\ref{fig:app_biosoccupation_template} in the appendix.
Across the four LLMs in our study, we find a consistent trend that, within each gender bucket along the horizontal axis, stereotypically feminine occupations lead to more positive dot products with the gender direction, whereas stereotypically masculine occupations cause the names in the same gender bucket to have more negative dot products.
This translates to higher predicted probability of the ``female'' token with the mention of a stereotypically feminine occupation, and vice versa.

We find that, except for \texttt{OLMo-7B}, LLMs tend to maintain their perceived gender associated with a name in comparison with $P_{\text{prior}}(\text{Female})$ for strongly feminine or masculine names, even with varying occupational contexts. Hence, the results for strongly feminine or masculine names are less affected by occupation than those for gender-ambiguous names.

\section{LLM Representations of Gender Influence Model Occupation Prediction}
\label{sec:downstream}

\begin{figure*}[t]
	\centering
        \begin{subfigure}[]{0.24\linewidth}
		\centering
		\includegraphics[width=\linewidth]{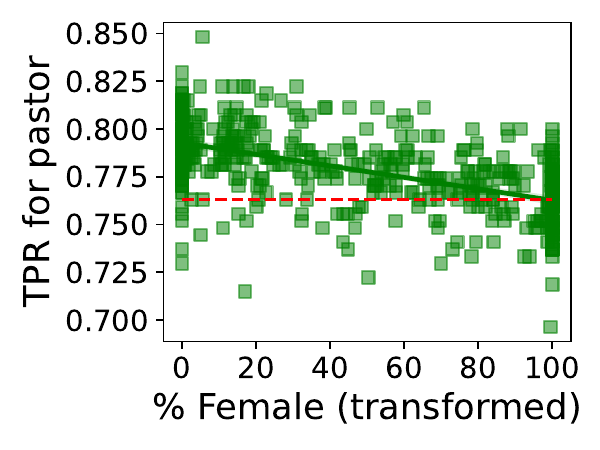}
		\caption{TPR(pastor) vs \% Female}
		\label{fig:pastor_bias}
	\end{subfigure}
	\hfill
        \begin{subfigure}[]{0.24\linewidth}
		\centering
        \includegraphics[width=\linewidth]{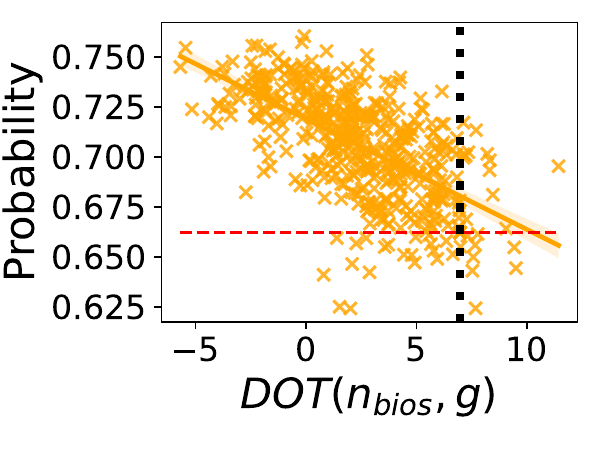}
		\caption{P(pastor) vs gender rep.}
		\label{fig:pastor_dot}
	\end{subfigure}
 	\begin{subfigure}[]{0.24\linewidth}
		\centering
		\includegraphics[width=\linewidth]{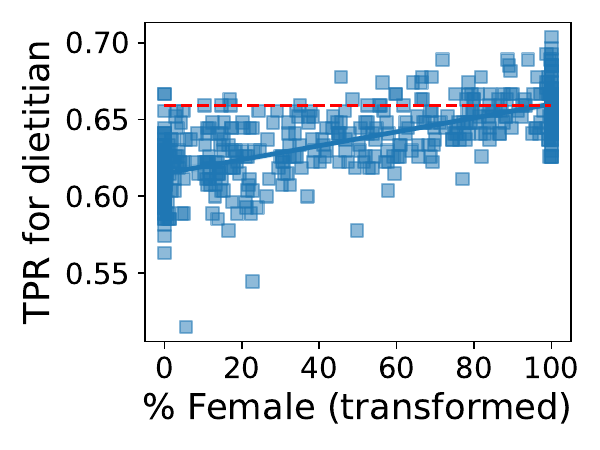}
		\caption{TPR(dietitian) vs \% Female}
		\label{fig:dietitian_bias}
	\end{subfigure}
    \hfill
    \begin{subfigure}[]{0.24\linewidth}
		\centering
        \includegraphics[width=\linewidth]{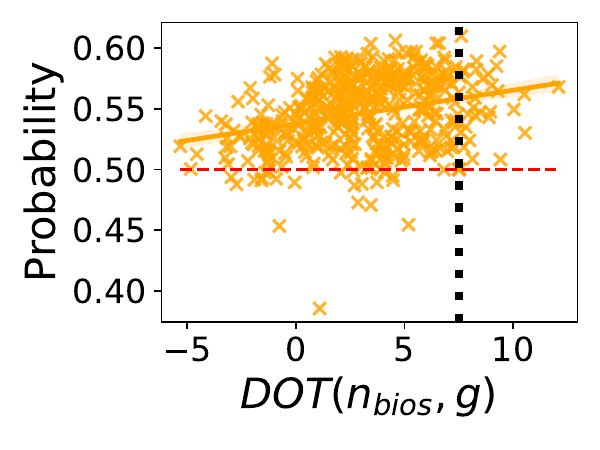}
		\caption{P(dietitian) vs gender rep.}
		\label{fig:dietitian_dot}
	\end{subfigure} 
    \caption{
    \textbf{(a) and (c):} \texttt{Llama-3.1-8B} shows higher TPR for masculine names in the male-dominated occupation ``pastor'' but lower TPR in the female-dominated occupation ``dietitian.'' The Pearson correlation in these plots represents the Bias Coefficients.
    \textbf{(b) and (d):} In \texttt{Llama-3.1-8B}, a more masculine first name increases the probability of ``pastor'' while feminine names have higher probabilities for ``dietitian,'' partly explaining the TPR gap.
    The Spearman correlation represents the Internal Coefficients (defined in~\cref{sec:bias_explanation}). Red dashed and black dotted lines show values when the first name is anonymized as ``X.''
    }
    
	\label{fig:bias_in_bios_corr}
\end{figure*}

\begin{table*}[]
\centering
\scriptsize
\resizebox{\linewidth}{!}{
    \begin{tabular}{p{0.9\linewidth}  p{0.03\linewidth}  p{0.04\linewidth}  p{0.03\linewidth}  p{0.1\linewidth}}
    \toprule
    \multicolumn{1}{c}{Biography}                                                     & \multicolumn{1}{c}{Label}  & \multicolumn{1}{c}{Name} & \multicolumn{1}{c}{\% Female} & \multicolumn{1}{c}{Prediction} \\ \midrule
    \multirow{6}{*}{
    \begin{minipage}[t]{\linewidth}
   
    Being a sports enthusiast, {[}NAME{]} was inspired by God to combine \_ passions of writing, sports, and Christ into a daily devotional that would encourage others to match their passion for Christ with their passion for their favorite team. {[}NAME{]}'s books, titled Daily Devotions for Die-Hard Fans and Daily Devotions for Die-Hard Kids, offer fans a unique mix of a true sports story connected to a daily reflection about God and their faith. The intent is to encourage the sports lover in a day-to-day walk with Christ through a devotion that is factual, Bible-based, and fun to read. Have fun. Have faith. Go God!
    \end{minipage}
    
    } & \multirow{6}{*}{pastor}    & Luis                & \multicolumn{1}{r}{0.53}                         & \textcolor{blue}{pastor}                  \\
    &                            & Logan                 & \multicolumn{1}{r}{7.37}                         & \textcolor{blue}{pastor}                     \\
    &                            & Jerre                 & \multicolumn{1}{r}{43.70}                         & \textcolor{blue}{pastor}                    \\
    &                            & Alejandra                  & \multicolumn{1}{r}{99.00}                         & \textcolor{red}{journalist}                     \\
    &                            & Khadijah                 & \multicolumn{1}{r}{99.90}                         & \textcolor{red}{journalist}                     \\
      &                            & Margarete                & \multicolumn{1}{r}{100.00}                        & \textcolor{red}{journalist}                     \\ \midrule
    \multirow{6}{*}{
    \begin{minipage}[t]{\linewidth} {[}NAME{]} lives in Long Island, NY where {[}NAME{]}'s works in foodservice and corporate wellness while also managing a virtual nutrition coaching practice. {[}NAME{]} specializes in intuitive eating and health at every size with a focus on sports nutrition. {[}NAME{]}'s blogs at KHNutrition.com where \_ loves to share about food, fitness and more recently, \_ journey with pregnancy and becoming a new mom.
    \end{minipage}
    }                                                                                                                                                   & \multirow{6}{*}{dietitian}   & Duc                      & \multicolumn{1}{r}{0.00}                        & \textcolor{red}{personal trainer}               \\
      &                            & Hunter                   & \multicolumn{1}{r}{5.02}                          & \textcolor{red}{personal trainer}               \\
      &                            & Dakota                   & \multicolumn{1}{r}{29.73}                         & \textcolor{red}{personal trainer}               \\ 
      &                            & Ivory                   & \multicolumn{1}{r}{59.32}                          & \textcolor{blue}{dietitian}               \\
      &                            & Laquinta                   & \multicolumn{1}{r}{97.82}                       &  \textcolor{blue}{dietitian}               \\
      &                            & Bonnie                   & \multicolumn{1}{r}{98.78}                         &  \textcolor{blue}{dietitian}               \\ \bottomrule
    \end{tabular} 
    }
    \caption{
    Example predictions from \texttt{Llama-3.1-8B} for two biographies with different substitutions of first names replacing the ``[NAME]'' placeholder for pastor (a male-dominated occupation) and dietitian (a female-dominated occupation).
    The model tends to make incorrect predictions when the perceived gender of a first name contradicts the stereotypical gender associated with the occupation. Correct predictions are \textcolor{blue}{blue}, and incorrect ones are \textcolor{red}{red}.
    }
    \label{tab:incorrect_examples}
\end{table*}

We study the influence of LLM representations of gender in a downstream occupation prediction task.
We demonstrate that gender representations in first-name embeddings can indicate biased behavior in LLMs for the high-stakes task of occupation prediction, despite some inconsistencies.

\subsection{Biased Behavior of LLMs}
\label{sec:biased_llms}

To show that LLMs reinforce gender-occupation stereotypes, we conduct zero-shot prompting experiments to predict occupation from biographies. We pose this as a multi-class classification task over a set of predefined occupations. 
We compute occupation probabilities by applying softmax to the logits of occupation tokens and consider the highest probability as the LLM's prediction.

\begin{prompt}[title={Prompt \thetcbcounter: Occupation Prediction}, label=prompt:occupation_prediction]
\scriptsize
Read the description about \{NAME\} below and predict their occupation.\\
\{BIO\}\\
What's \{NAME\}'s occupation? Output an occupation only. No preambles. No explanations.
\end{prompt}

\paragraph{Setup} We use the Bias in Bios dataset~\citep{dearteaga-2019-biasinbios} that contains biographies of individuals across $28$ occupations. For each occupation, we randomly sample $135$ female and male biographies respectively.
As all names and gendered pronouns are redacted in the biographies, we replace the name placeholder with one of the $470$ first names from~\cref{sec:gender_dir_eval} and use the prompt template~\ref{prompt:occupation_prediction} to ask an LLM to predict the occupation.
This prompt choice follows~\citet{sancheti-etal-2024-influence}, who conducted prompt tuning before selecting their final design for a similar classification task.
In total, we prompt each LLM $3,553,200$ times to assess biases in occupation prediction.

\paragraph{Bias Coefficient} 
Following~\citet{dearteaga-2019-biasinbios}, we use true positive rate (TPR) to measure prediction gaps between female and male names. For each occupation, we compute TPR of a name by substituting the name into the same set of $270$ biographies. 
The scatter plots of TPR often show a linear trend between the femininity of a name and the model's performance, revealing a TPR gap between feminine and masculine names. The bias coefficient, defined as the Pearson correlation of scatter points, reflects the strength of the linear relationship between the associated femininity of a name and the model's performance. A value near $0$ with a large $p$ suggests that the stereotypical gender of a name does not correlate with model performance, showing similar performance across genders. A significantly positive value indicates a higher TPR for feminine names, and vice versa.

\begin{figure*}[t]
	\centering
        \begin{subfigure}[]{0.495\linewidth}
		\centering
		\includegraphics[width=\linewidth]{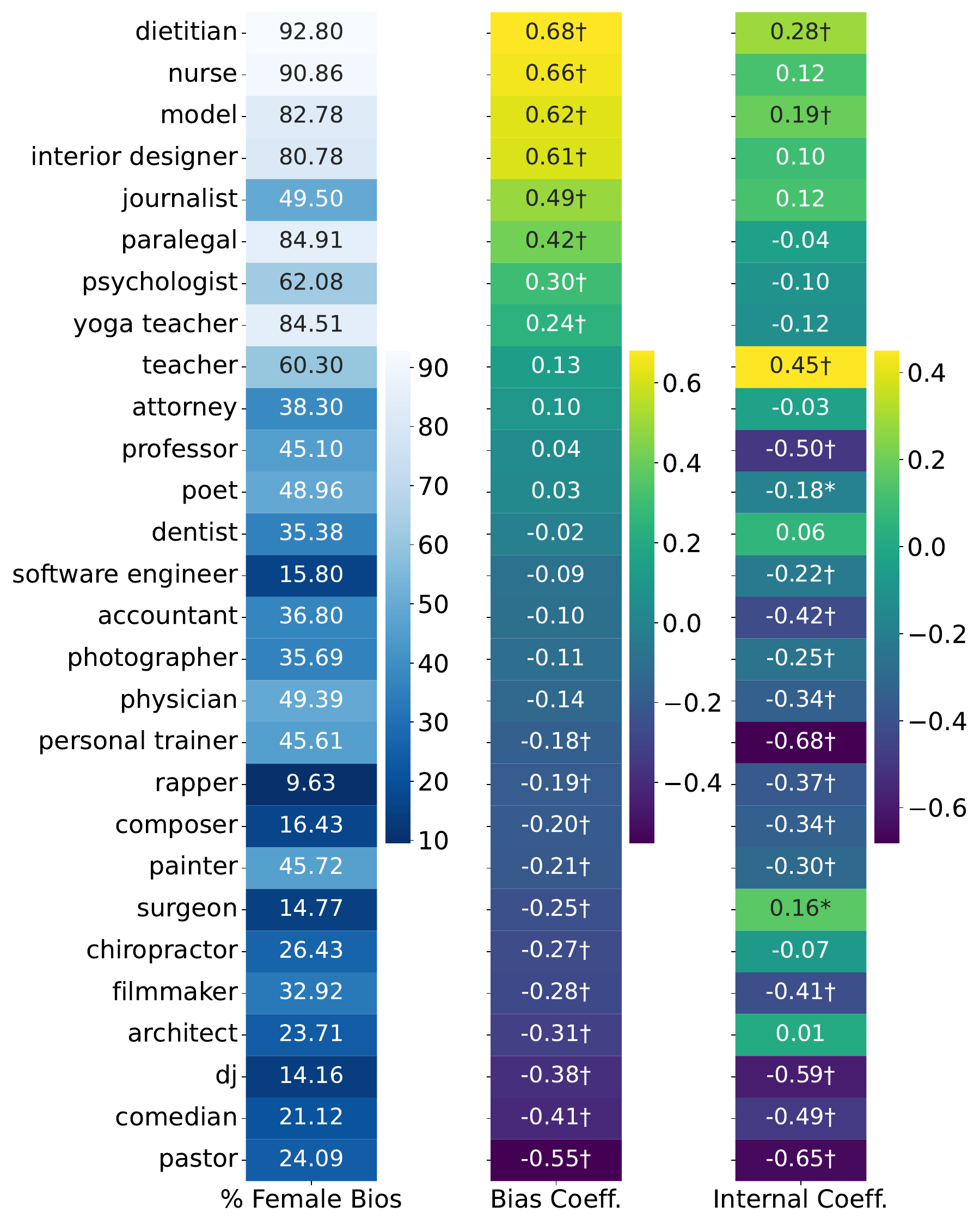}
		\caption{Llama-3.1-8B}
		\label{fig:heatmap_llama}
	\end{subfigure}
	\hfill
        \begin{subfigure}[]{0.495\linewidth}
		\centering
		\includegraphics[width=\linewidth]{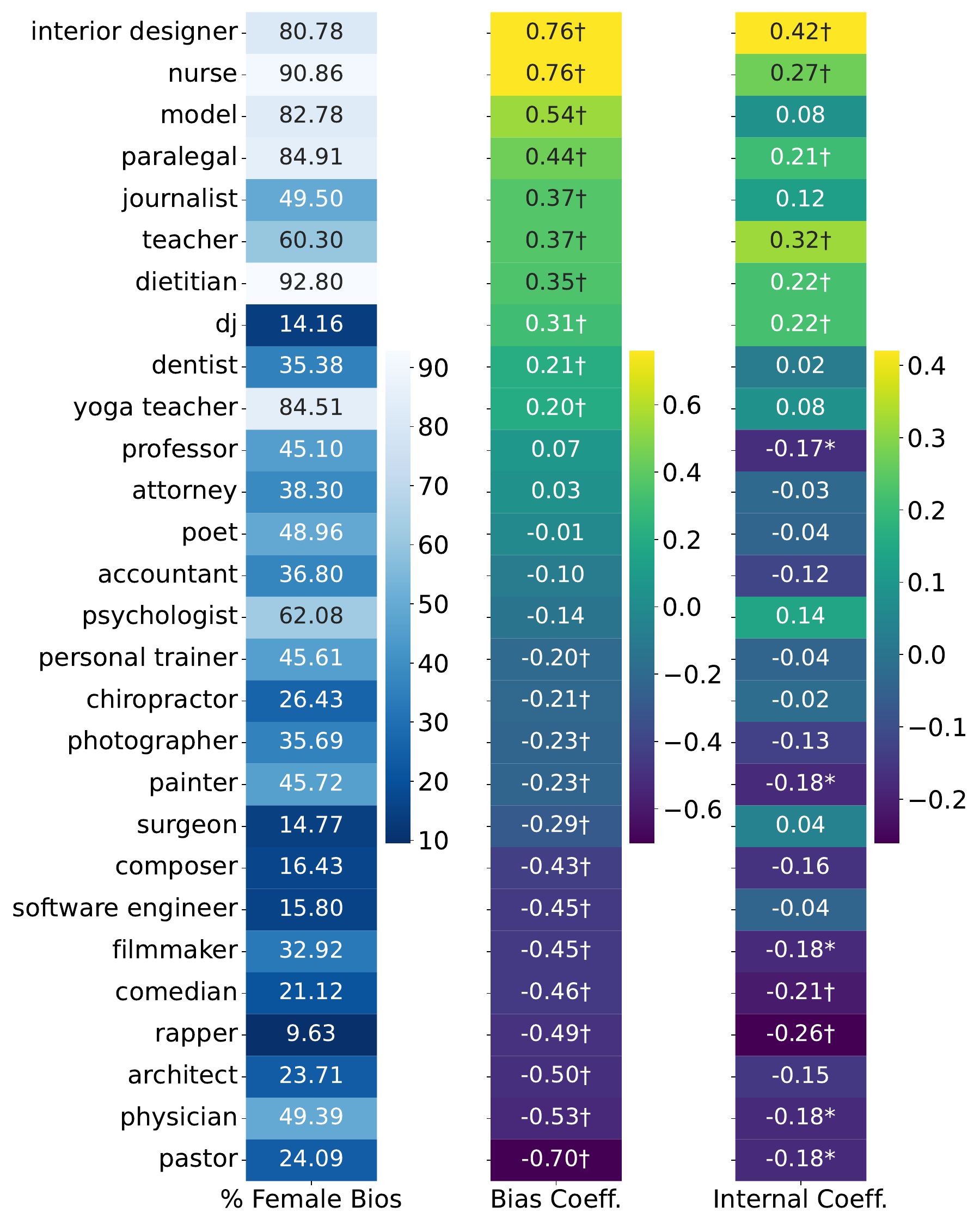}
		\caption{Mistral-7B}
		\label{fig:heatmap_mistral}
	\end{subfigure}
    \caption{Percentage of female biographies, Bias Coefficient (sorted in descending order), and the Spearman correlation between $DOT(\vec{n}_{\text{bios}, g})$ and the predicted probability for each occupation, defined as the ``Internal Coefficient.''
    The bias coefficient and internal coefficient are moderately correlated, despite some inconsistencies. 
    Notations: $\dagger$: $p < 0.001$. $*$: $p < 0.005$.
    All $p$ values are corrected using the Holm–Bonferroni method.
        }
	\label{fig:heatmap_dot_and_bias}
\end{figure*}

\paragraph{Gender Bias in Occupation Predictions} We report the results in Fig.~\ref{fig:bias_in_bios_corr}.
We observe a negative ($-0.55$) and positive ($0.68$) bias coefficient, respectively in Fig.~\ref{fig:pastor_bias} and Fig.~\ref{fig:dietitian_bias}, for \texttt{Llama-3.1-8B}.
Feminine names tend to receive higher TPR for ``dietitian'' ($92.80\%$ female) while masculine names generally have higher TPR for ``pastor'' ($24.09\%$ female).
Hence, the model achieves higher TPR when a first name's perceived gender aligns with the occupation's stereotypical gender, reinforcing gender-occupation stereotypes.
All biased occupation predictions are shown in Fig.~\ref{fig:heatmap_dot_and_bias} (middle columns) with $\dagger$, indicating a statistically significant Pearson correlation ($P<0.001$).
In \textsection\ref{sec:bias_explanation}, we investigate LLM's internal gender representations to offer potential explanations to the observed bias.

\paragraph{Examples of Predictions}
We present a few example predictions from \texttt{Llama-3.1-8B} in Table~\ref{tab:incorrect_examples}.
Given the same biography with different first-name substitutions, the model tends to misclassify strongly feminine names for the male-dominated occupation pastor and masculine names for the female-dominated occupation dietitian.
While these are anecdotal examples from two biographies and a small subset of first names, the next section examines the broader trend between the internal gender representation of first names and the model's extrinsic biased behavior.

\subsection{Gender Representation Partially Explains the Biased Behavior} \label{sec:bias_explanation}
We analyze internal first-name representations and intermediate LLMs outputs (occupation token logits) to explore their correlation. If gender representation correlates with occupation logits, it may partly explain the observed extrinsic model biases.

We retrieve the contextualized embedding of the last occurrence of a first name in the prompt and compute its dot product with $\vec{g}$, yielding DOT$(\vec{n}_{\text{bios}}, g)$, averaged across $270$ contexts for each occupation. We also compute the averaged predicted probability of the ground-truth occupation token. We plot these two variables in Fig.~\ref{fig:pastor_dot} and Fig.~\ref{fig:dietitian_dot} for ``pastor'' and ``dietitian.'' The figures show that a more masculine representation (more negative dot product) increases the probability for ``pastor'' but decreases it for ``dietitian,'' 
consistent with the model's behavior in Fig.~\ref{fig:pastor_bias} and Fig.~\ref{fig:dietitian_bias}. Although the correlation between DOT$(\vec{n}_{\text{bios}}, g)$ and $P(\text{occupation})$ appears linear,
the ranking is more important, so we compute Spearman's correlation as the \textbf{Internal Coefficient}.

In Fig.~\ref{fig:heatmap_dot_and_bias}, we show the internal coefficients for all occupations in two LLMs, with similar results for the other two LLMs in~\cref{sec:app_additional}. The two coefficients are moderately correlated (Spearman's correlation is $0.61$ for \texttt{Llama-3.1-8B} and $0.76$ for \texttt{Mistral-7B}, both $p<0.001$), 
suggesting that what we observe internally in the model has some correlation with the extrinsic biased behavior.

However, in Fig.~\ref{fig:heatmap_dot_and_bias}, the internal coefficient sometimes fails to capture biased extrinsic behavior (e.g., ``nurse'' and ``journalist'') and occasionally produces false positives (e.g., ``physician'' and accountant''). 
These limitations highlight the challenges of using gender representation for bias prediction, echoing earlier findings that intrinsic and extrinsic metrics do not necessarily align with each other~\cite{goldfarb-tarrant-etal-2021-intrinsic, cao-etal-2022-intrinsic}.

\section{Conclusion}
In this paper, we approximate and rigorously evaluate a gender direction in state-of-the-art LLMs. Using a validated gender direction, we analyze the femininity of first-name embeddings in both controlled and real-world contexts. We find that the gender representation of first names interacts with stereotypical occupations in context, sometimes revealing model bias in downstream tasks. However, the noisy correlation between the model's internal gender representation in first-name embeddings and its extrinsic biased behavior underlines the need for more robust methods to detect bias using internal gender representations.

\section*{Limitations}
\paragraph{Underrepresentation of Gender Identities}
Following the seminal work of~\citet{bolukbasi2016man}, we approximate a female-male gender direction in the embedding space of LLMs. While we have included gender-ambiguous first names in our study, the gender direction approximation may underrepresent gender identities beyond this binary definition.
We acknowledge this limitation and leave the inclusion of additional gender identities in embedding analysis for future work.

\paragraph{Limited Coverage of Demographic Identities}
Our selection of first names from the available data sources (SSA and~\citet{rosenman2023race}) is limited to two genders and four races/ethnicities within the U.S. context. Unfortunately, many demographic identities could not be included due to insufficient data availability. Collecting additional first names that represent other genders, races, and ethnicities is essential for a more comprehensive study of first-name representations in LLMs, although it remains a challenging task.
Various other demographic attributes, such as age, nationality, and religion, could also be studied~\cite{parrish-etal-2022-bbq, hou2025language}. However, we find it infeasible to obtain sufficient data sources to use first names as proxies for these demographic attributes.

\paragraph{Under-Exploration of Model Size}
The size of a language model may significantly affect its performance and the extent of biases it exhibits~\cite{tal-etal-2022-fewer, srivastava2022beyond}, influencing both representational and allocational harm~\cite{barocas2017problem, crawford2017troublewithbias, blodgett-etal-2020-language}. However, due to resource constraints, our experiments -- totaling over 12 million input prompts -- were conducted exclusively on smaller-sized LLMs. While these findings provide some insights, they leave open the question of whether similar patterns and biases persist or intensify in larger models.
Future research could address this gap by investigating the relationship between model size and bias, particularly to determine if the trends observed here scale consistently across larger models. 

\paragraph{Lack of Mitigation Solutions}
While our results highlight biased behavior of LLMs in predicting occupations from biographies toward first names associated with different genders, we do not propose immediate solutions to mitigate these biases. Instead, our paper focuses on the interpretability of the models' internal gender representations in first-name embeddings and their correlation with the models' extrinsic behavior. Nonetheless, bias mitigation remains a critical research direction.

\section*{Ethics Statement}
\paragraph{Perceived Gender and Self-Identifications}
The task of predicting a person's gender from their name may raise ethical concerns. 
Gender identity is defined by the HRC Foundation\footnote{\url{https://www.hrc.org/resources/}} as ``one's innermost concept of self as male, female, a blend of both, or neither – how individuals perceive themselves and what they call themselves. One's gender identity can be the same or different from their sex assigned at birth.''
While we introduce the binary gender classification task as a test of the gender direction approximation, we strongly discourage using predicted gender to oversimplify the diverse gender identities associated with a name.
Our further analysis reveals the model's representation of the perceived gender associated with a first name in various occupational contexts, but this perception may differ from an individual's self-identified gender.
The discrepancy between perceived and self-identified gender can lead to disrespect and misunderstandings. While there is no easy or universal solution to the over-generalization of gender in first-name embeddings from LLMs, we argue that we must strive to build inclusive technologies that minimize such harm. Notable efforts include those by~\citet{cao-daume-iii-2020-toward, baumler-rudinger-2022-recognition, piergentili-etal-2023-gender, piergentili-etal-2024-enhancing, bartl-leavy-2024-showgirls}, among others. Our paper contributes to understanding internal gender representations in LLMs, paving the way for the development of gender-inclusive language technologies.

\paragraph{Gender-Occupation Stereotype}
Due to imbalances in the gender breakdown of many occupations, the corpora on which models are trained can inherit these gender-occupation biases, leading to the development of gender-occupation stereotypes in the models' downstream behavior.
Our observations show that LLMs continue to rely on the over-generalization of gender-occupation correlations when making predictions. Ongoing efforts are needed to address this biased behavior in LLMs.

\section*{Acknowledgments}
We thank the anonymous reviewers for their insightful comments, which helped improve our paper.
Rachel Rudinger and Haozhe An are supported by NSF CAREER Award No. 2339746. Any opinions, findings, and conclusions or recommendations expressed in this material are those of the author(s) and do not necessarily reflect the views of the National Science Foundation.

\bibliography{custom, anthology}

\appendix

\section{Gendered Words}
\label{sec:app_gendered_words}
We display the lists of gendered words and random words in Table~\ref{tab:def_words}. These words are used to approximate a female-male gender direction and a random direction in LLMs, respectively.

\begin{table*}[t]
\centering
\resizebox{\linewidth}{!}{ 
    \begin{tabular}{@{}l|llllllllll@{}}
    \toprule
    Female   & she  & her      & woman     & herself & daughter & mother & gal   & girl     & female & Mary     \\
    Male     & he   & his      & man       & himself & son      & father & guy   & boy      & male   & John     \\ \midrule
    Random 1 & book & sun      & ice       & tree    & flower   & river  & house & dog      & car    & mountain \\
    Random 2 & vase & elephant & xylophone & jungle  & umbrella & pencil & kite  & notebook & guitar & zebra    \\ \bottomrule
    \end{tabular}           
}
    \caption{Gendered words for finding a female-male gender direction and a random direction in the embedding space.
        }
	\label{tab:def_words}
\end{table*}

\section{First Names}
\label{sec:app_first_names}

In Table~\ref{tab:name_distribution}, we present the breakdown of the demographic statistics for the first names used in our study.
In addition, the $24$ names (which is a subset of the sampled names) we used to obtain the contexts for averaged first-name embedding computation are:
\begin{itemize}
    \item Female: Carie (White), Marybeth (White), Darci (White),  Khadijah (Black), Yashica (Black), Tamiko (Black), Miguelina (Hispanic), Agueda (Hispanic), Betzaida (Hispanic), Quynh (Asian), Huong (Asian), Thuy (Asian);
    \item Male: Jerad (White), Zoltan (White), Benjamen (White),  Cedric (Black), Trayvon (Black), Demarco (Black), Osvaldo (Hispanic), Luis (Hispanic), Rigoberto (Hispanic), Dong (Asian), Huy (Asian), Khoa (Asian).
\end{itemize}

\begin{table*}[t]
\centering
\resizebox{\linewidth}{!}{ 
    \begin{tabular}{l|rrrrrrrrrr|r}
    \toprule
    \% Female         & 0-2 & 2-5 & 5-10 & 10-25 & 25-50 & 50-75 & 75-90 & 90-95 & 95-98 & 98-100 & Total \\ \midrule
    White    & 30  & 21  & 8    & 5     & 6     & 10    & 9     & 13    & 17    & 30     & 149   \\
    Black    & 30  & 9   & 14   & 18    & 6     & 12    & 8     & 10    & 27    & 30     & 164   \\
    Hispanic & 30  & 1   & 0    & 1     & 0     & 0     & 1     & 0     & 4     & 30     & 67    \\
    Asian    & 14  & 2   & 5    & 4     & 11    & 7     & 11    & 3     & 6     & 27     & 90    \\ \midrule
    Total    & 104 & 33  & 27   & 28    & 23    & 29    & 29    & 26    & 54    & 117    & 470   \\ \bottomrule
    \end{tabular}
}
\caption{The distribution of sampled first names by percentage of female from real-world statistics for each race/ethnicity in our study.
        }
	\label{tab:name_distribution}
\end{table*}

\section{Additional Results}
\label{sec:app_additional}

\begin{figure*}[t]
	\centering
 	\begin{subfigure}[]{0.49\linewidth}
		\centering
		\includegraphics[width=\linewidth]{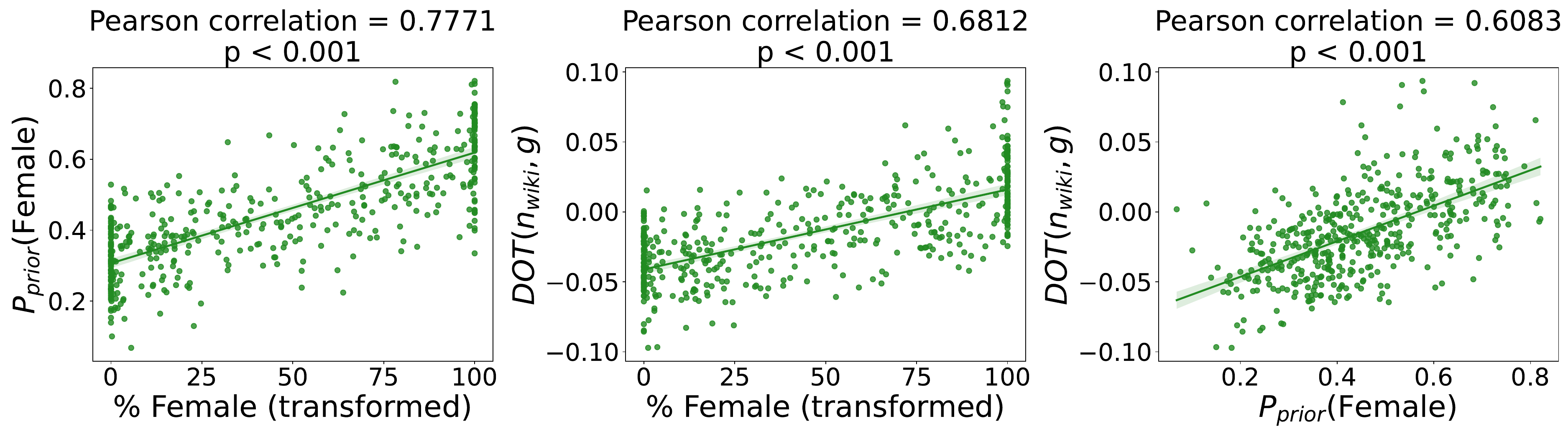}
		\caption{OLMo-7B}
		\label{fig:pearsonr_olmo}
	\end{subfigure}
    \hfill
    \begin{subfigure}[]{0.49\linewidth}
		\centering
		\includegraphics[width=\linewidth]{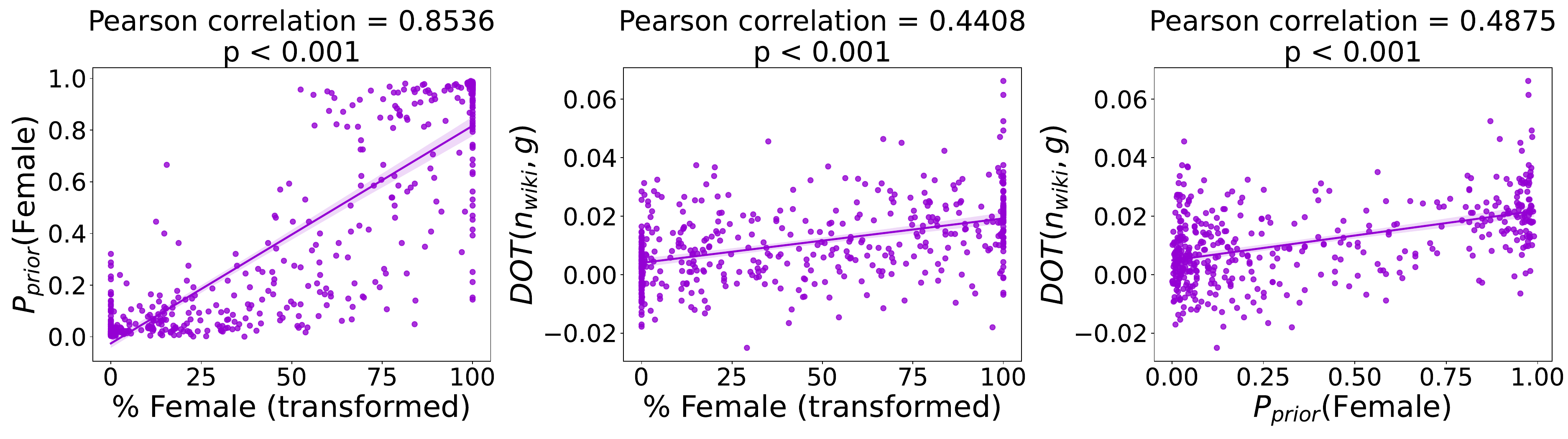}
		\caption{Phi-3.5-mini}
		\label{fig:pearsonr_phi}
	\end{subfigure} 
    \caption{Additional results for \texttt{OLMo-7B} and \texttt{Phi-3.5-mini}. Scatter plot between each pair of the three variables studied in~\cref{sec:align_rep_and_world} and their Pearson correlation.}
	\label{fig:app_pearsonr_3variables}
\end{figure*}

\begin{figure*}[t]
 	\begin{subfigure}[]{0.49\linewidth}
		\centering
		\includegraphics[width=\linewidth]{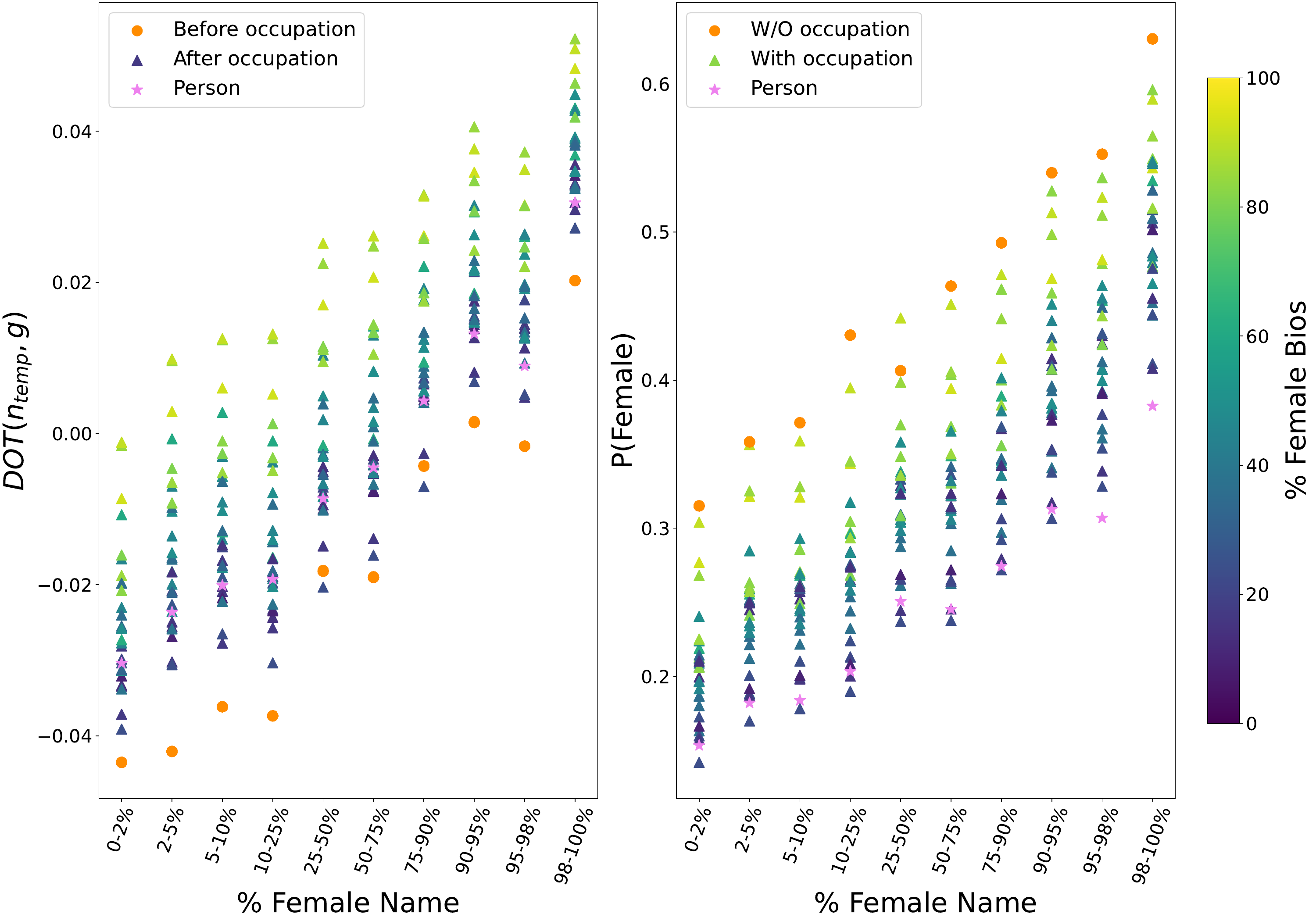}
		\caption{OLMo-7B}
		\label{fig:template_olmo}
	\end{subfigure}
    \hfill
    \begin{subfigure}[]{0.49\linewidth}
		\centering
		\includegraphics[width=\linewidth]{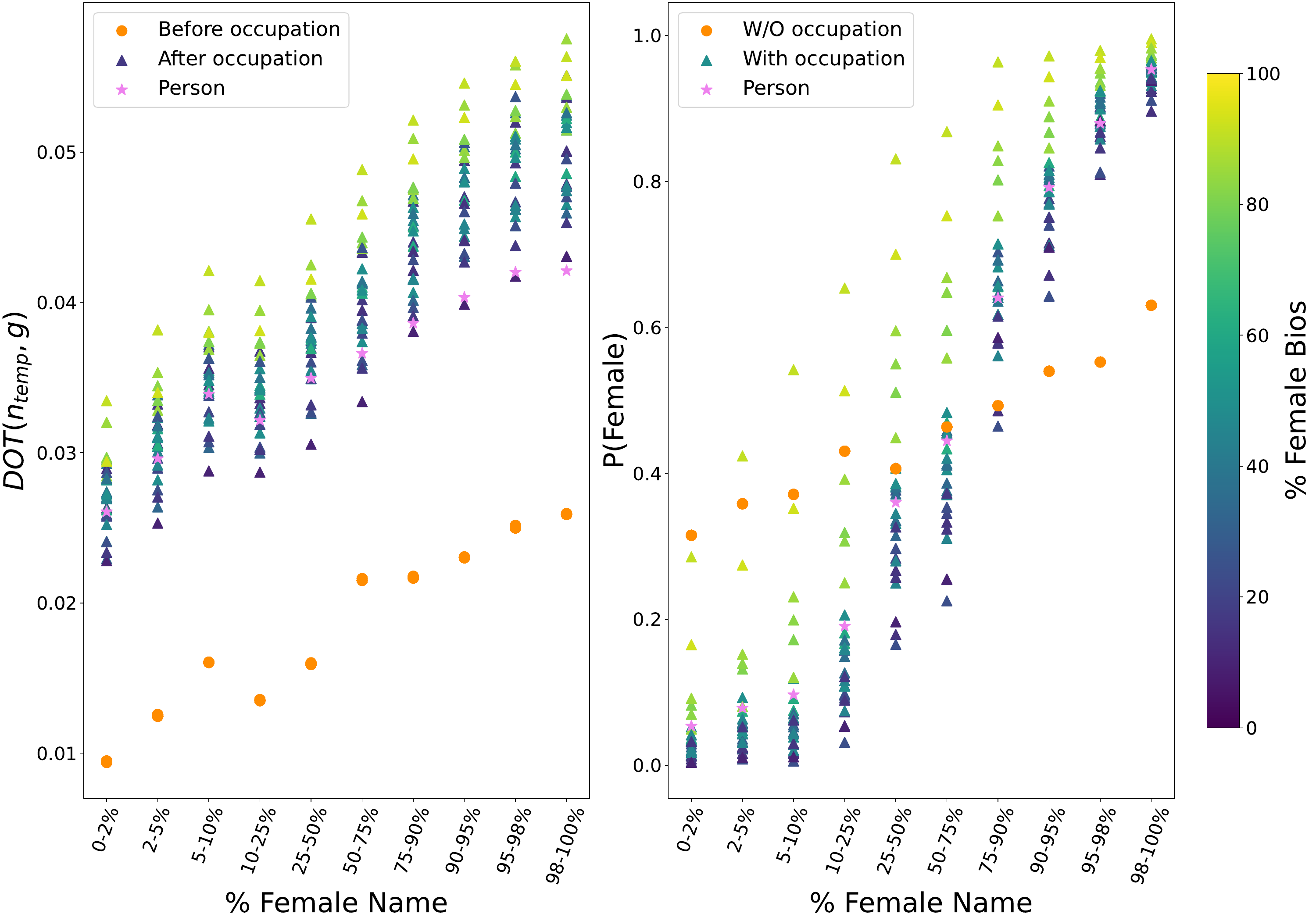}
		\caption{Phi-3.5-mini}
		\label{fig:template_phi}
	\end{subfigure} 
    \caption{Additional results for \texttt{OLMo-7B} and \texttt{Phi-3.5-mini}. (Left of each subfigure) Change of the dot product between the name embedding from a template sentence $\vec{n}_{\text{temp}}$ and the gender direction $\vec{g}$ before and after the mention of an occupation.
    (Right of each subfigure) Change of the output probability of the token ``female'' with and without mentioning an occupation.}
	\label{fig:app_biosoccupation_template}
\end{figure*}

We present the additional experimental results for \texttt{OLMo-7B} and \texttt{Phi-3.5-mini} in Fig.~\ref{fig:app_pearsonr_3variables} and Fig.~\ref{fig:app_biosoccupation_template}. The experiment details are described in~\cref{sec:align_rep_and_world} and~\cref{sec:rep_and_context} respectively.

\begin{figure*}[t]
	\centering
        \begin{subfigure}[]{0.49\linewidth}
		\centering
		\includegraphics[width=\linewidth]{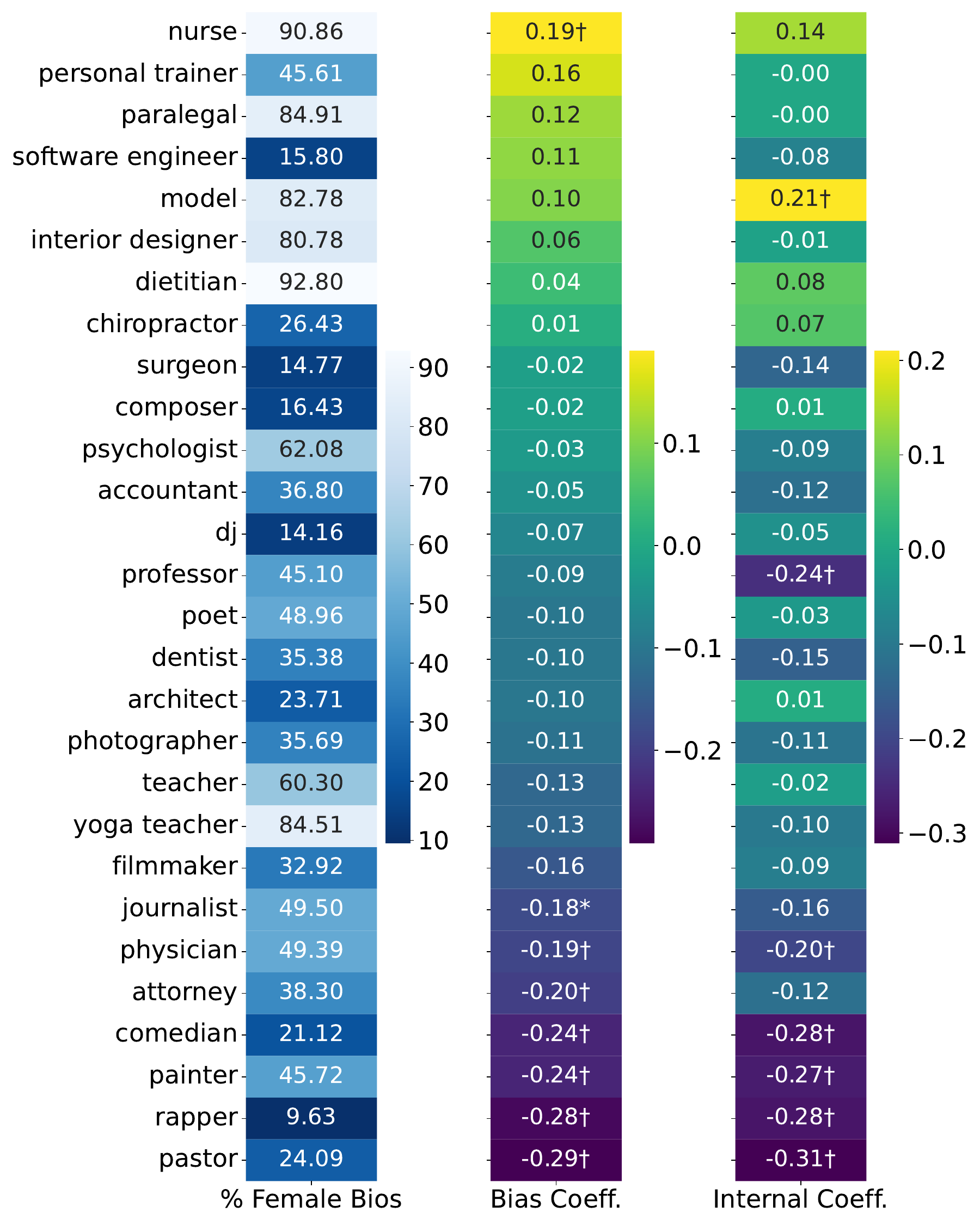}
		\caption{OLMo-7B}
		\label{fig:heatmap_olmo}
	\end{subfigure}
	\hfill
        \begin{subfigure}[]{0.49\linewidth}
		\centering
		\includegraphics[width=\linewidth]{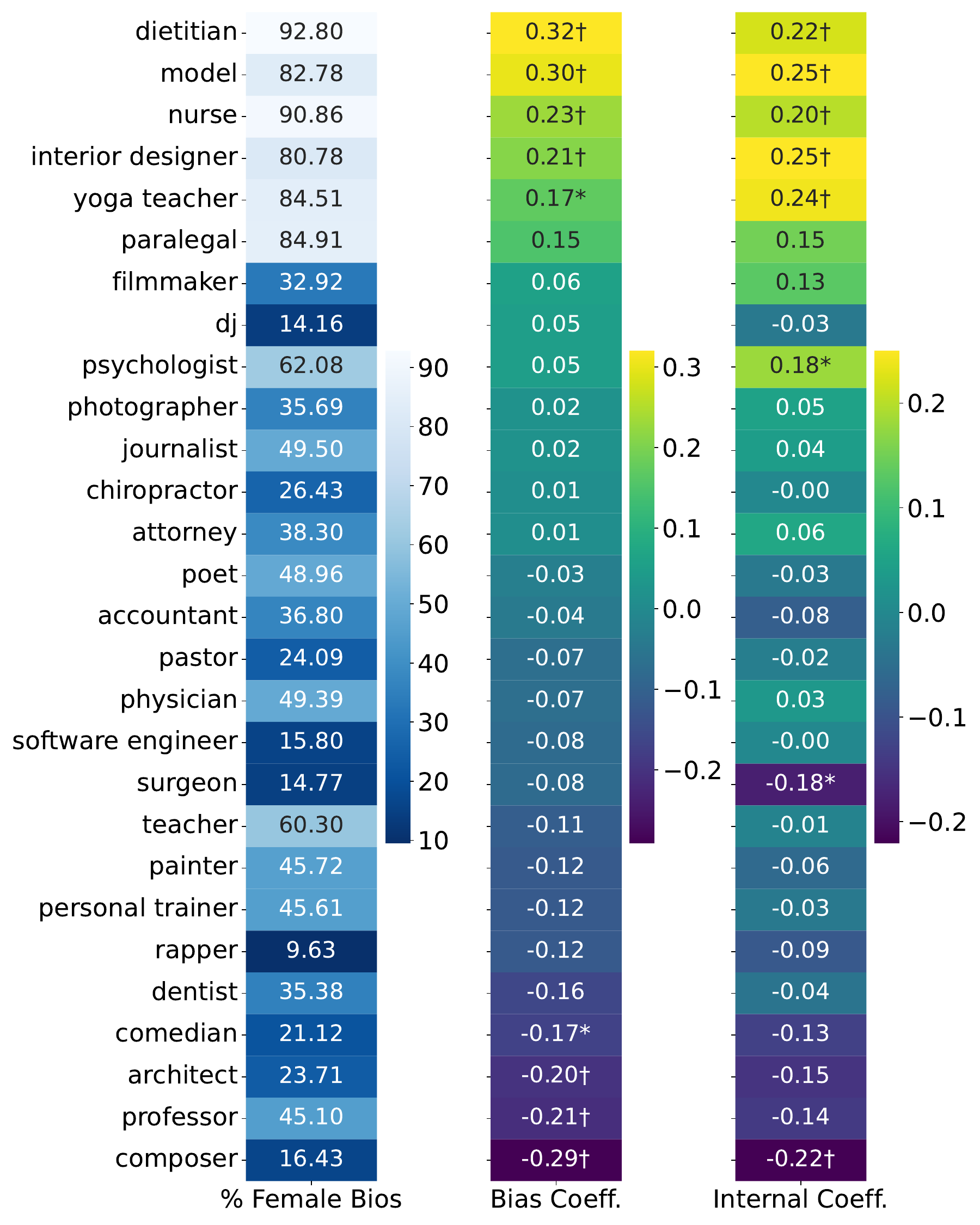}
		\caption{Phi-3.5-mini}
		\label{fig:heatmap_phi}
	\end{subfigure}
    \caption{Additional results for \texttt{OLMo-7B} and \texttt{Phi-3.5-min} from the experiments described in~\cref{sec:bias_explanation}. The plots show the percentage of female biographies, Bias Coefficient (sorted in descending order), and the Spearman correlation between $DOT(\vec{n}_{\text{bios}, g})$ and the predicted probability for each occupation, defined as the ``Internal Coefficient.''
    Notations: $\dagger$: $p < 0.001$. $*$: $p < 0.005$.
    All $p$ values are corrected using the Holm–Bonferroni method.
        }
	\label{fig:app_heatmap_dot_and_bias}
\end{figure*}

For the comparison between Bias Coefficient and Internal Coefficient (\cref{sec:bias_explanation}), we show the results for \texttt{OLMo-7B} and \texttt{Phi-3.5-mini} in Fig.~\ref{fig:app_heatmap_dot_and_bias}. 
Consistent with the results discussed in~\cref{sec:bias_explanation}, the internal coefficients also correlate with the bias coefficients for these two LLMs, showing a Spearman's correlation of $0.86$ for \texttt{OLMo-7B} and $0.90$ for \texttt{Phi-3.5-mini}, with $p < 0.001$ in both cases.

\section{Models}
We list the source of each model that has been used in this paper. 
All model usage is consistent with their respective intended use.

\begin{itemize}
    \item \texttt{Llama-3.1-8B-Instruct}\\Model is available  at \url{https://huggingface.co/meta-llama/Llama-3.1-8B-Instruct}. Llama 3.1 is intended for commercial and research use in multiple languages with a Llama 3.1 Community License Agreement.)
    
    \item \texttt{Mistral-7B-Instruct-v0.3}\\  Model is available  at \url{https://huggingface.co/mistralai/Mistral-7B-Instruct-v0.3}. Mistral-7B comes with an Apache 2.0 License that allows redistribution of the work or derivative works.

    \item \texttt{OLMo-7B-0724-hf}\\Model is available  at \url{https://huggingface.co/allenai/OLMo-7B-0724-hf}. OLMo-7B also comes with an Apache 2.0 License that allows redistribution of the work or derivative works.

    \item \texttt{Phi-3.5-mini-instruct}\\Model is available  at \url{https://huggingface.co/microsoft/Phi-3.5-mini-instruct}. The model is intended for commercial and research use in multiple languages with an MIT license. 
\end{itemize}

Each model is run on an NVIDIA RTX A5000 GPU. Due to the large scale of our empirical study, which includes $470$ first names and biographies across $28$ occupations, the total computational time amounts to approximately $2,000$ GPU hours.

\end{document}